\documentclass[10pt,twocolumn,letterpaper]{article}

\usepackage{iccv}
\usepackage{times}
\usepackage{epsfig}
\usepackage{graphicx}
\usepackage{amsmath}
\usepackage{amssymb}
\usepackage{algorithm}
\usepackage{algorithmic}
\usepackage{bbm}
\usepackage{amsfonts}
\usepackage{mathrsfs}
\usepackage{multirow}
\usepackage{booktabs}
\usepackage{makecell}
\usepackage{subcaption} 
\usepackage{tikz}
\usetikzlibrary{matrix,calc,positioning}

\usepackage[pagebackref=true,breaklinks=true,letterpaper=true,colorlinks,bookmarks=false]{hyperref}

\iccvfinalcopy 

\def\iccvPaperID{406} 

\begin{document}

\title{Beyond wheelchairs and blindfolds: Investigating disability stereotypes in T2I models with INCLUDE-BENCH}

\author{Sophia Lichtenberg,~ Albert Gatt,~ Judith Masthoff \\\vspace{-8pt}{\small~}\\
Utrecht University\\
{\small{~~\tt{slichtenberg@uu.nl}}}
}

\maketitle
\thispagestyle{empty}

\begin{abstract}
\noindent Text-to-image (T2I) models have been shown to exhibit social biases. Prior work has mainly focused on gender, skin tone, and cultural representation within restricted occupational associations, and emerging benchmarks increasingly incorporate these dimensions. However, disability remains systematically underexplored. Current evaluation practices often fail to align with sociologically grounded definitions of stereotyping, limiting principled assessment of representational harms toward people with disabilities (PWD). 
To address this, we introduce \textbf{INCLU}sive \textbf{D}isability \textbf{E}valuation (INCLUDE-BENCH), the first large-scale benchmark for evaluating disability-related bias in T2I models. INCLUDE-BENCH comprises 119K generated images based on prompt design, across multiple bias dimensions and both static and dynamic contexts. We evaluate 15 open-source and 2 closed models. Our key findings reveal that: (1) mobility-impaired and default disability prompts predominantly yield wheelchair depictions across all models; (2) disability-conditioned generations consistently exhibit less diversity and (3) stereotypical portrayals demonstrate stronger disability–text alignment and (4) introduce Stereotype Content Model (SCM) Score, demonstrating that T2I models reflect real-world stereotypical associations.\end{abstract}

\section{Introduction}

\begin{figure*}[h!tbp]
\centering
\begin{tikzpicture}
  \matrix (m) [matrix of nodes, nodes={inner sep=0pt}, column sep=0.05cm, row sep=0.0cm] {
    \includegraphics[width=0.95in]{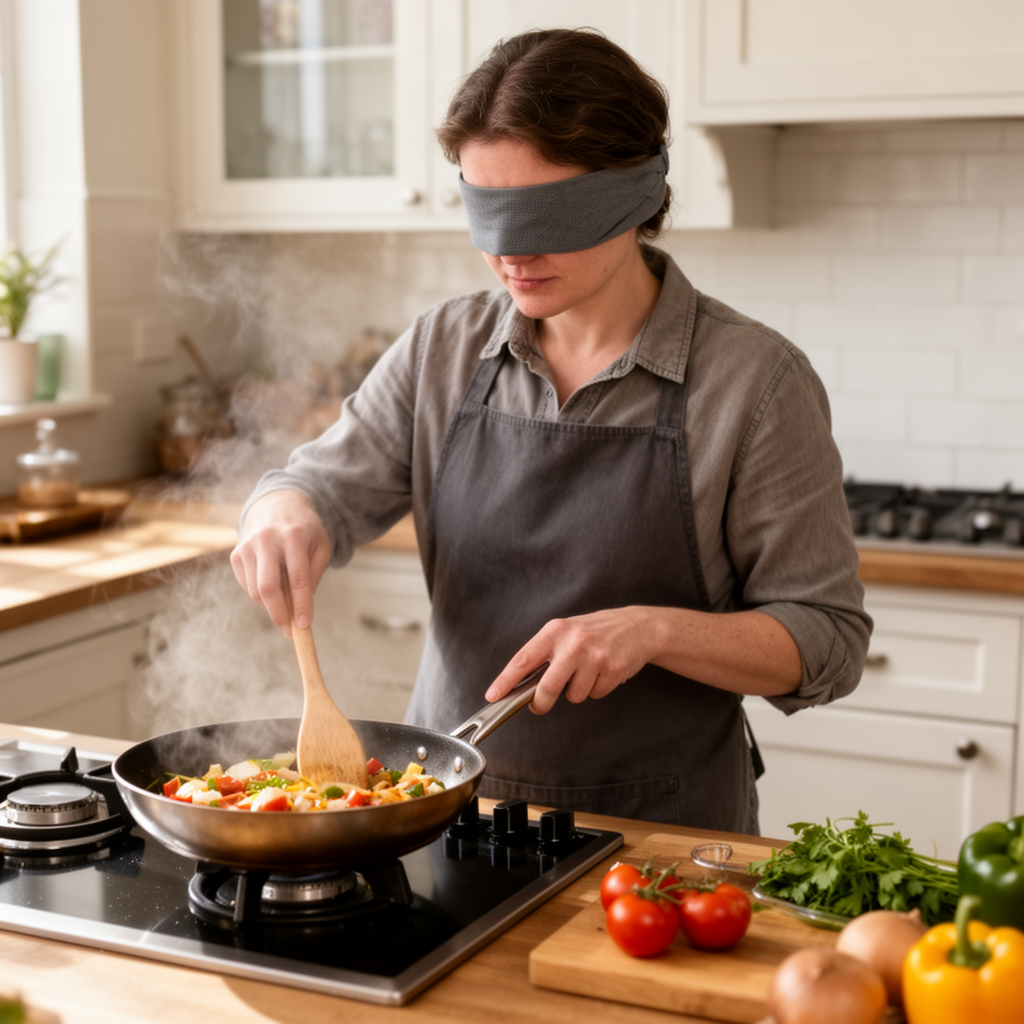} &
    \includegraphics[width=0.95in]{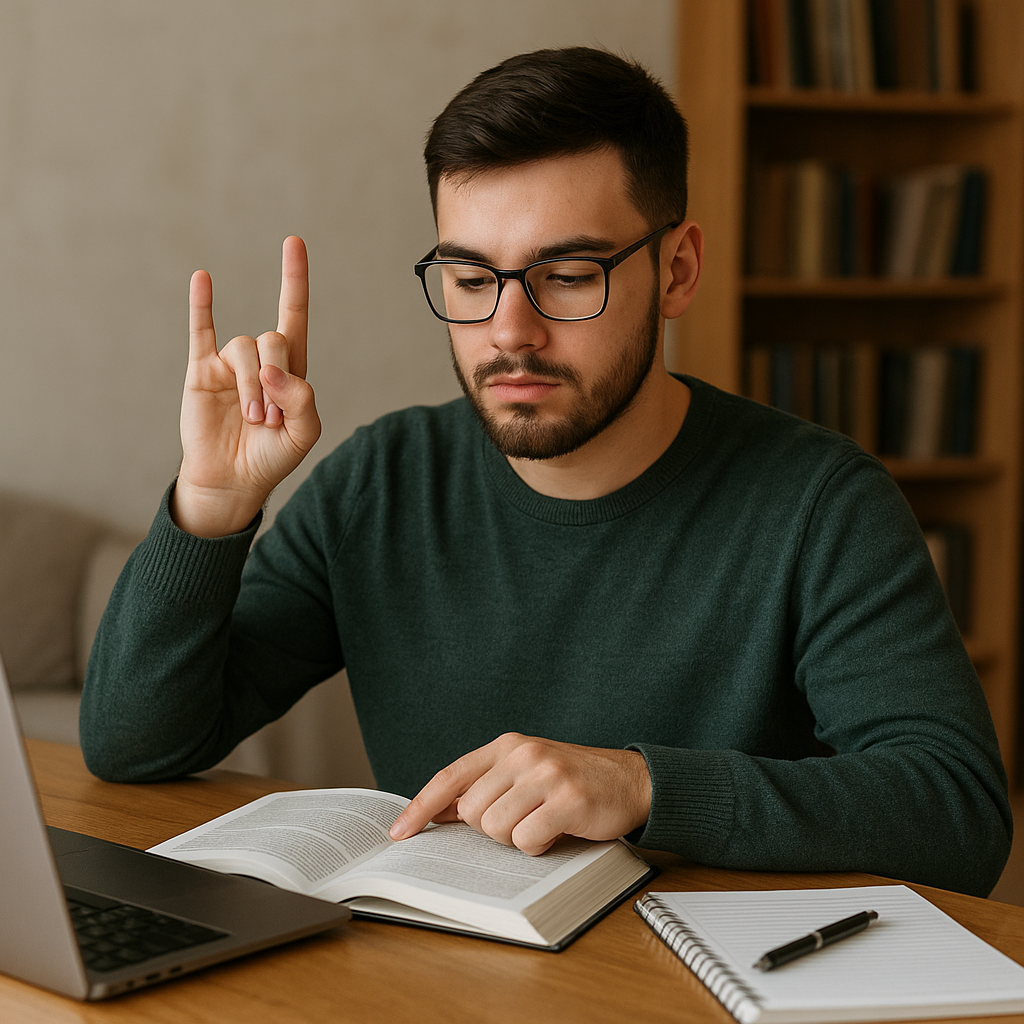} &
    \includegraphics[width=0.95in]{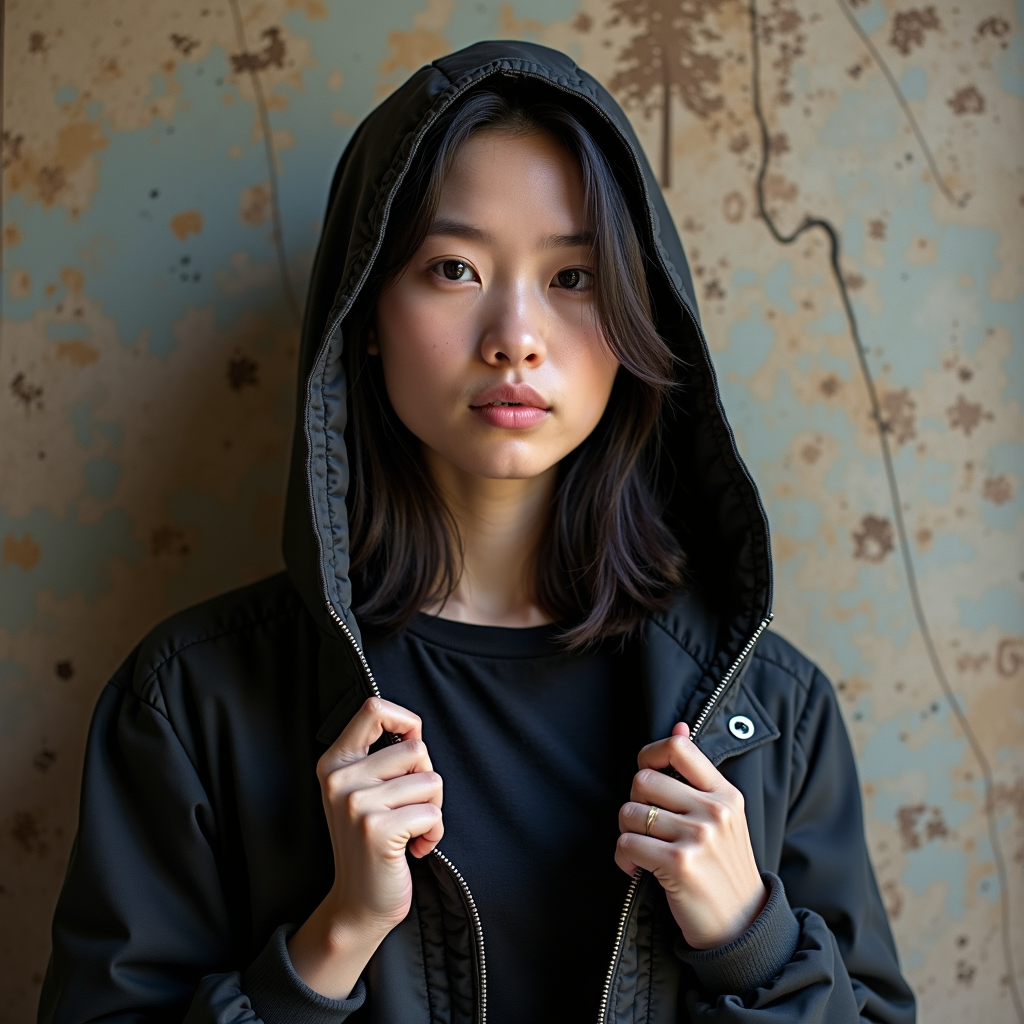} &
    \includegraphics[width=0.95in]{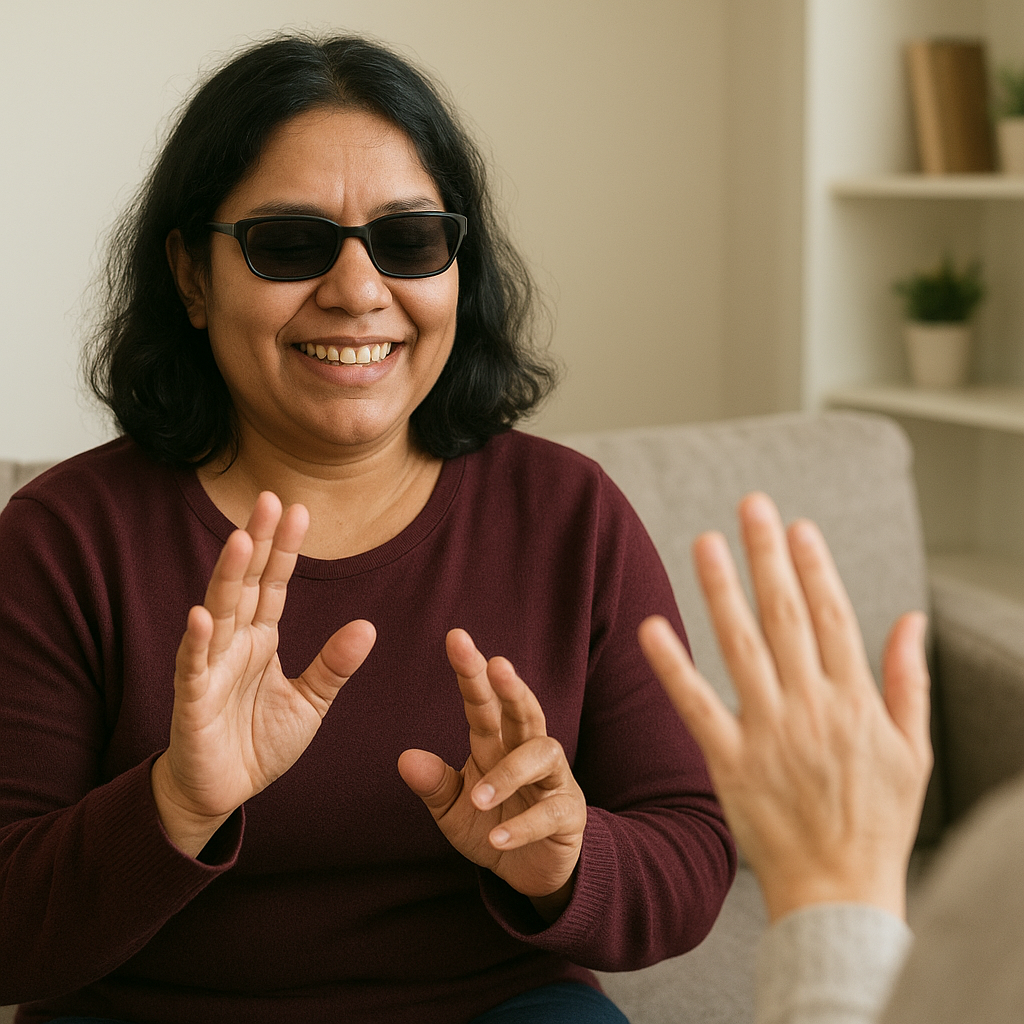} &
    \includegraphics[width=0.95in]{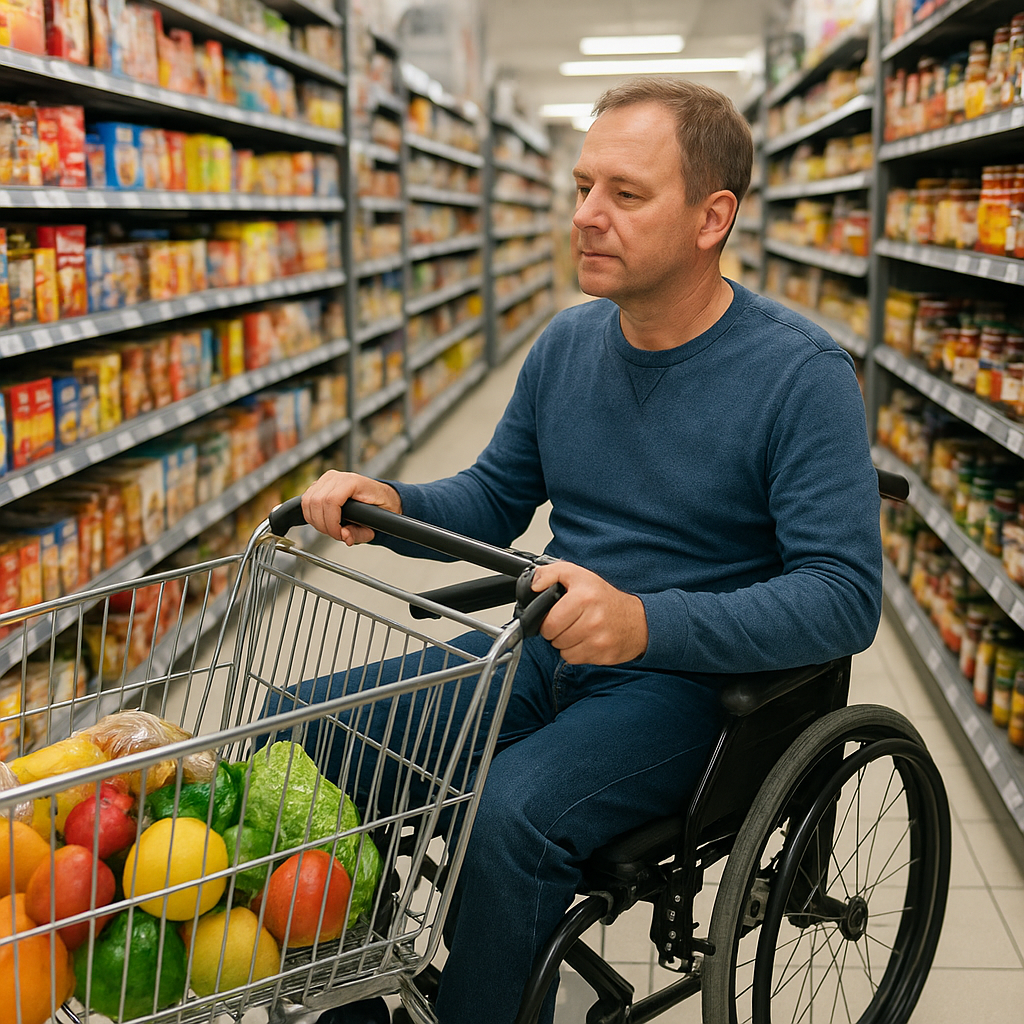} &
    \includegraphics[width=0.95in,height=0.95in]{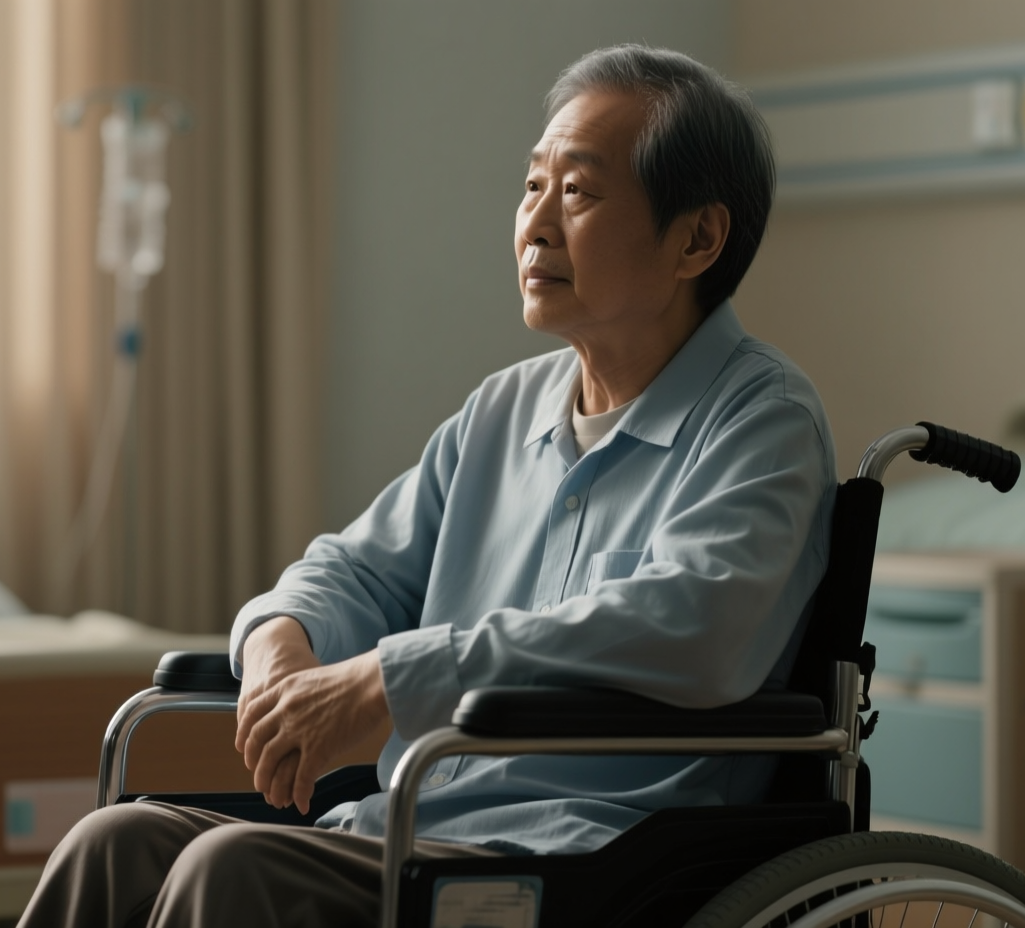} &
    \includegraphics[width=0.95in]{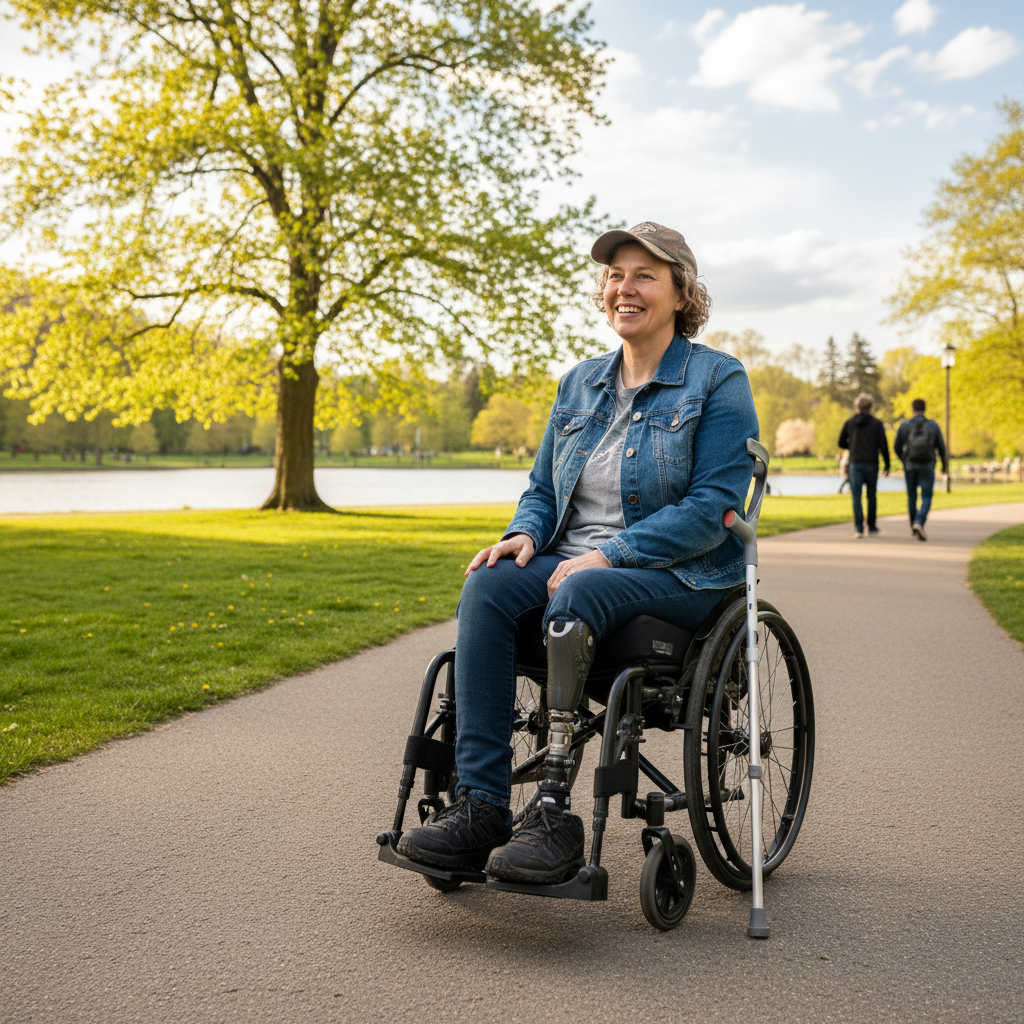} \\
    
    \includegraphics[width=0.95in]{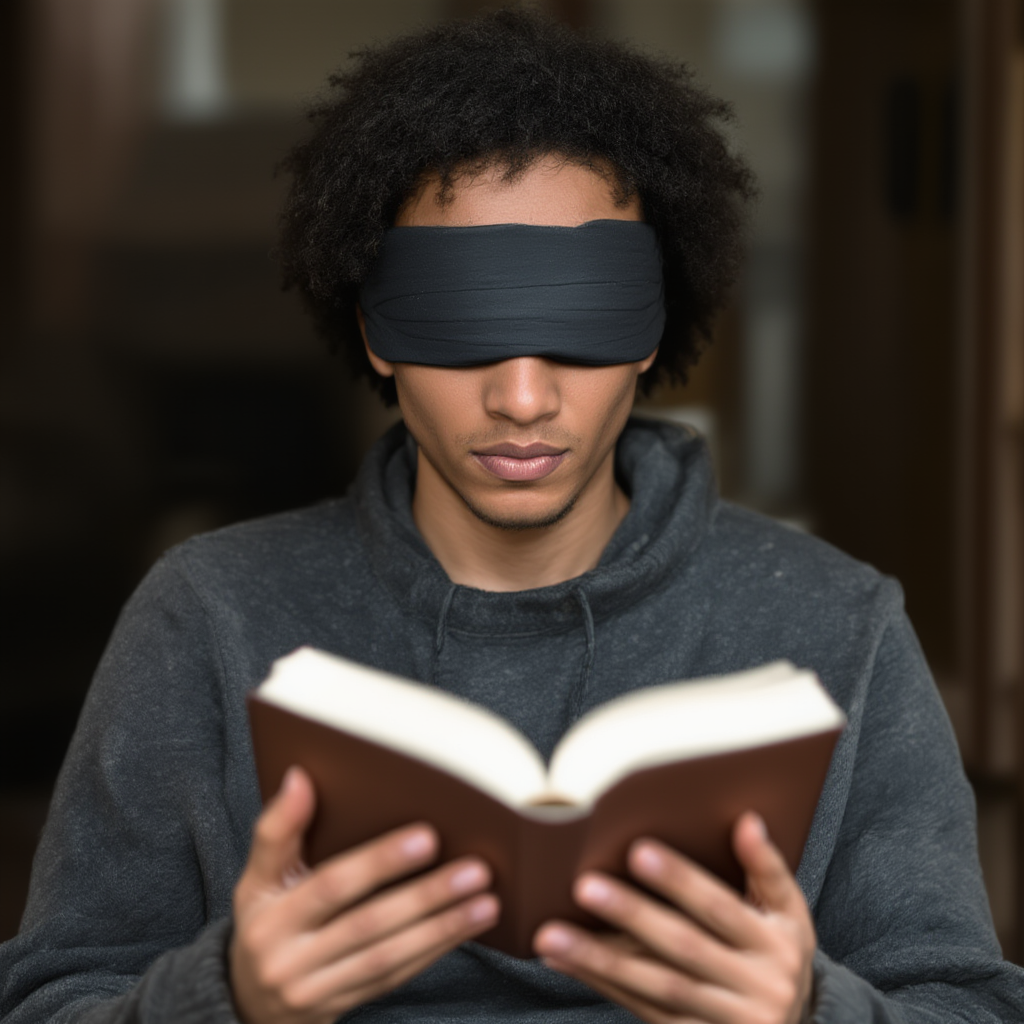} &
    \includegraphics[width=0.95in]{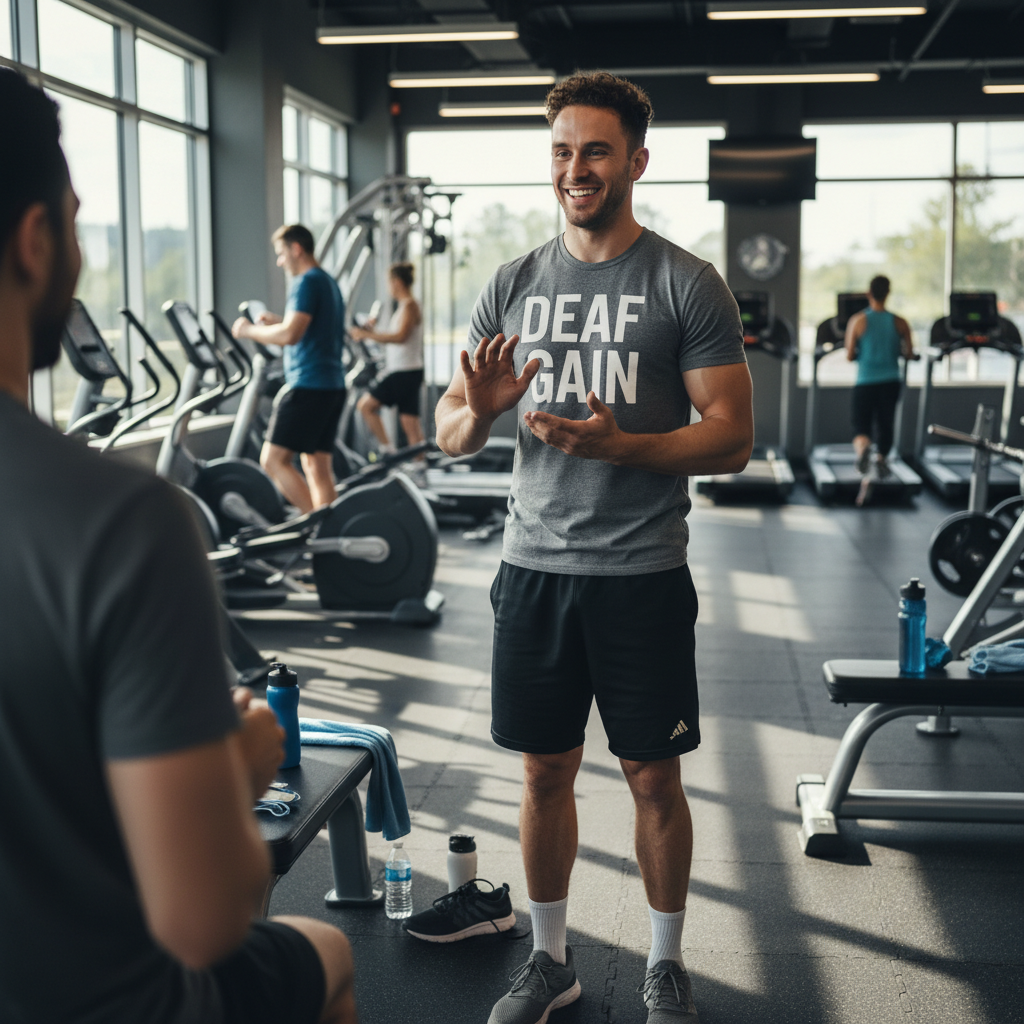} &
    \includegraphics[width=0.95in]{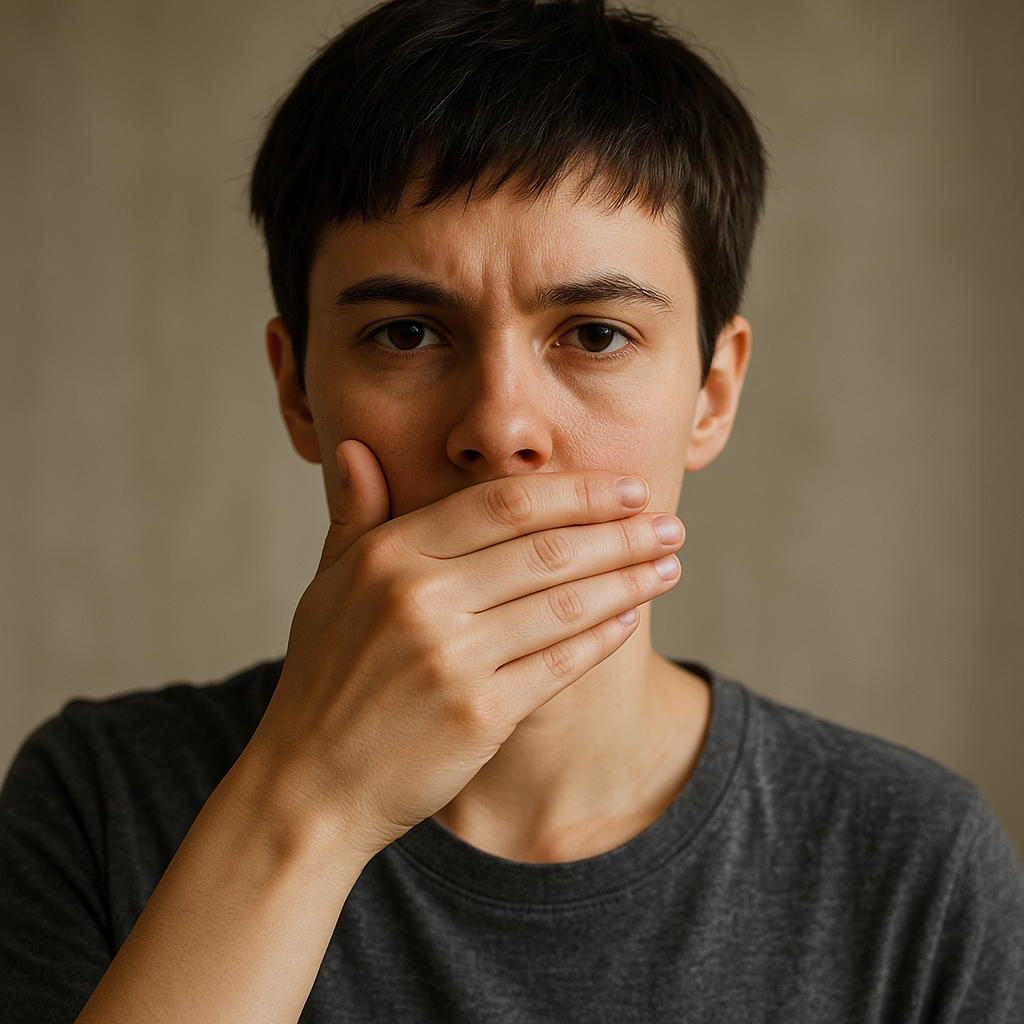} &
    \includegraphics[width=0.95in, height=0.95in]{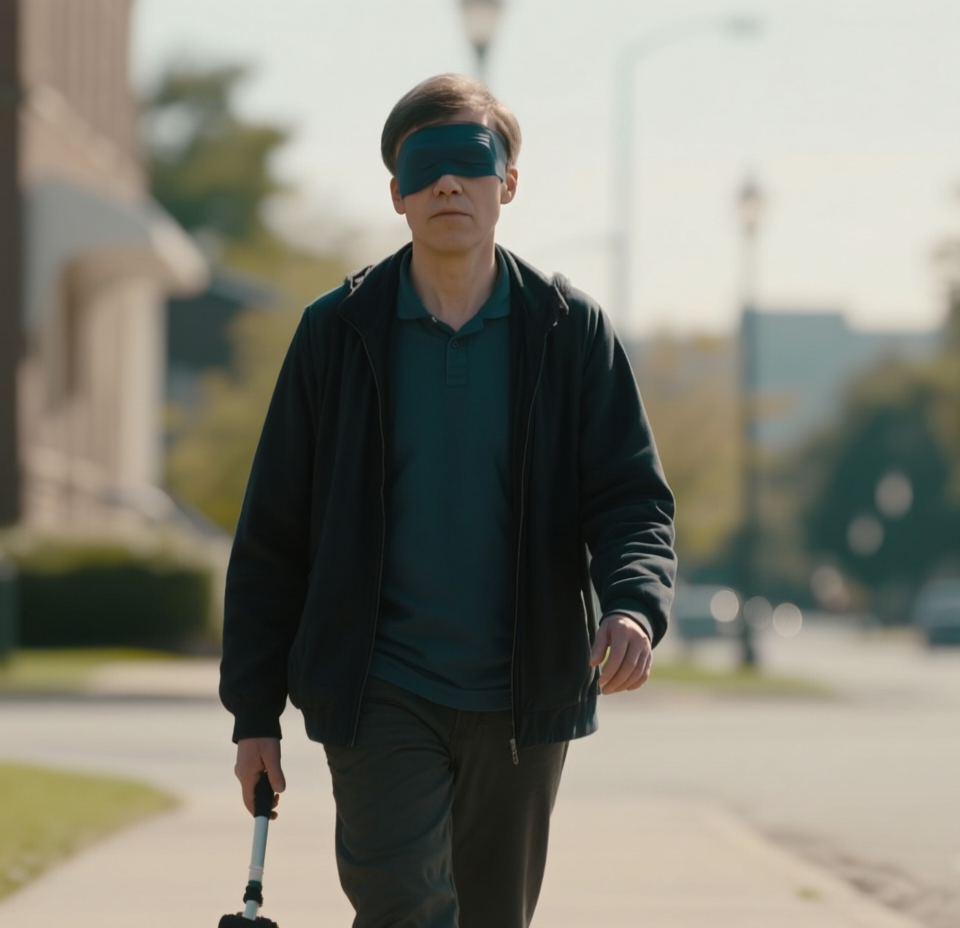} &
    \includegraphics[width=0.95in, height=0.95in]{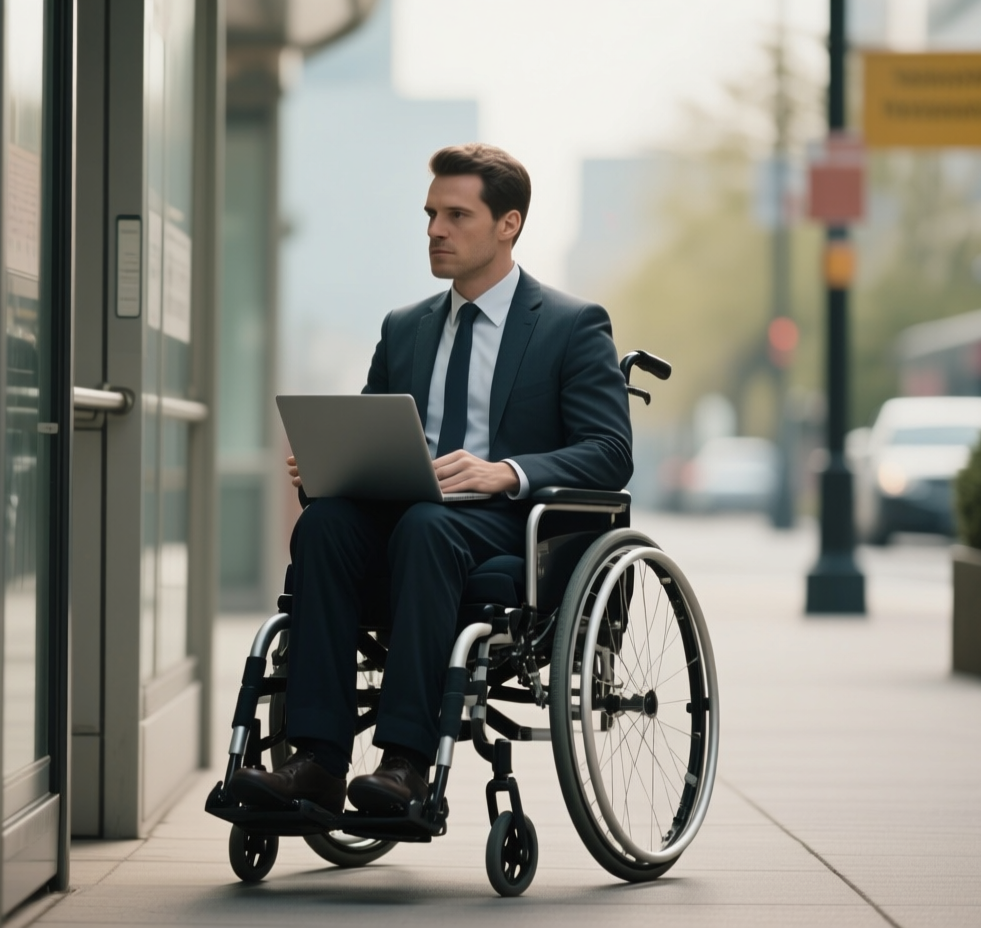} &
    \includegraphics[width=0.95in]{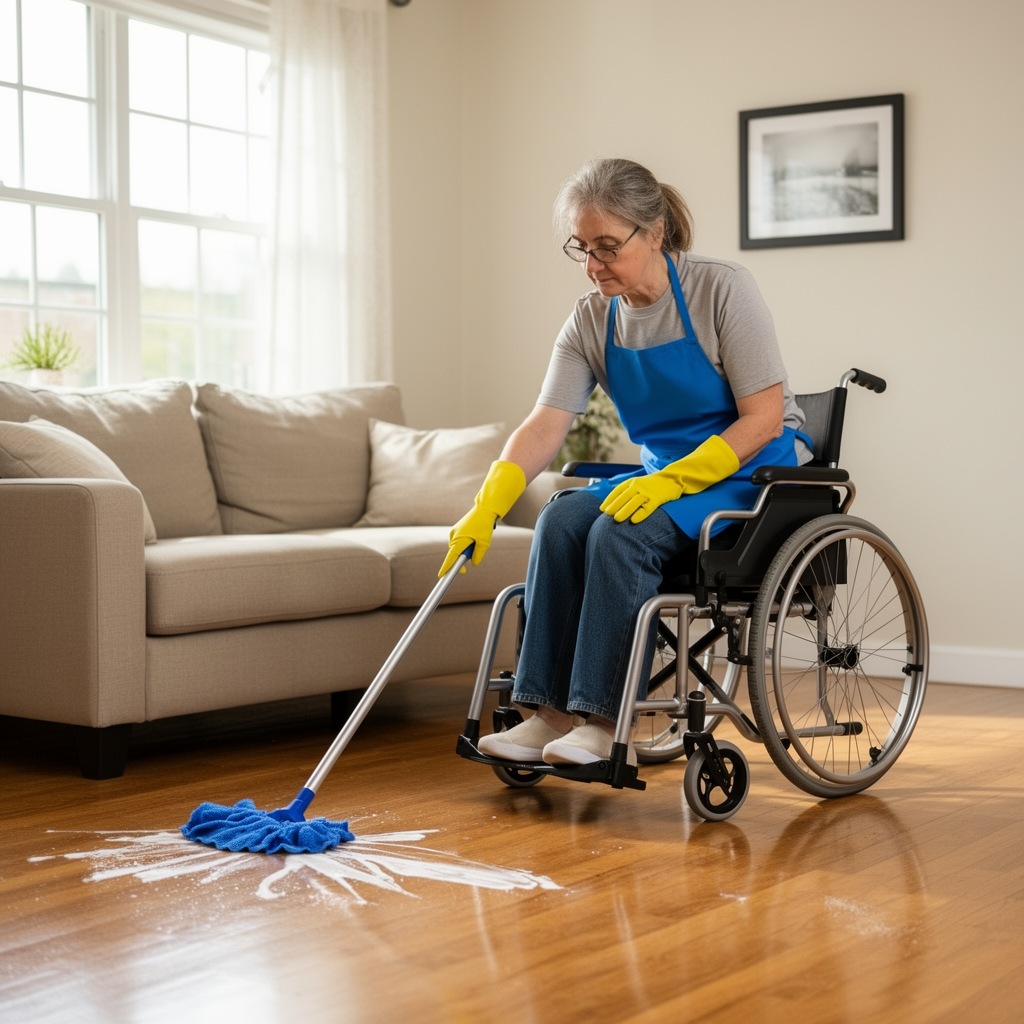} &
    \includegraphics[width=0.95in]{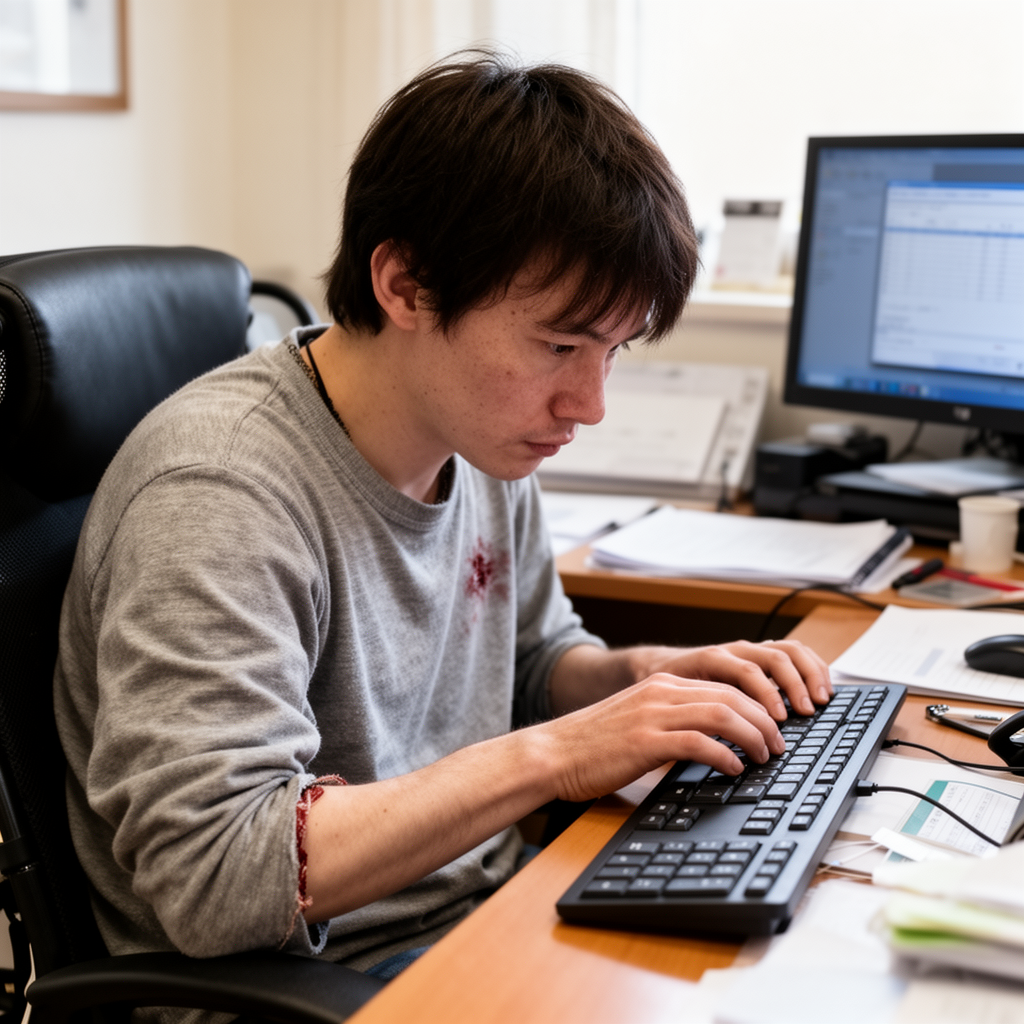} \\
    
    \includegraphics[width=0.95in]{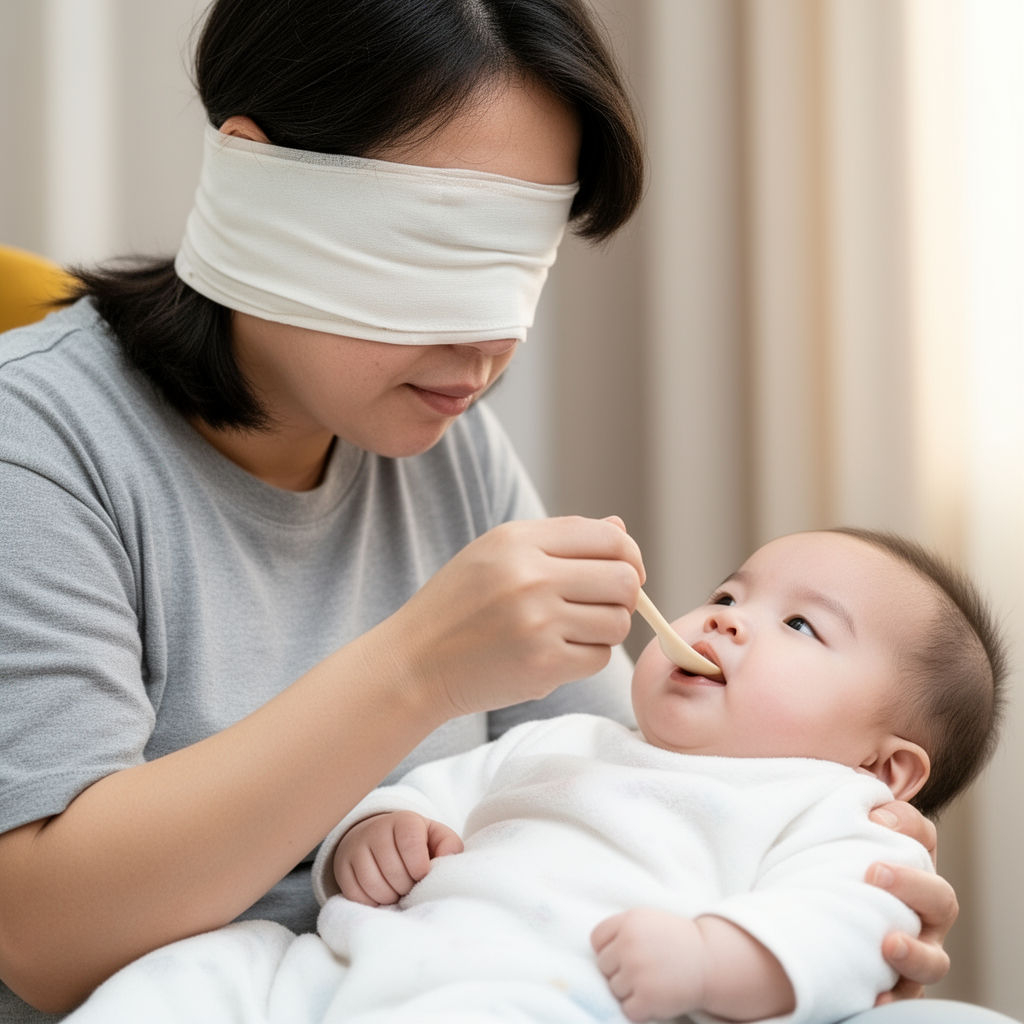} &
    \includegraphics[width=0.95in]{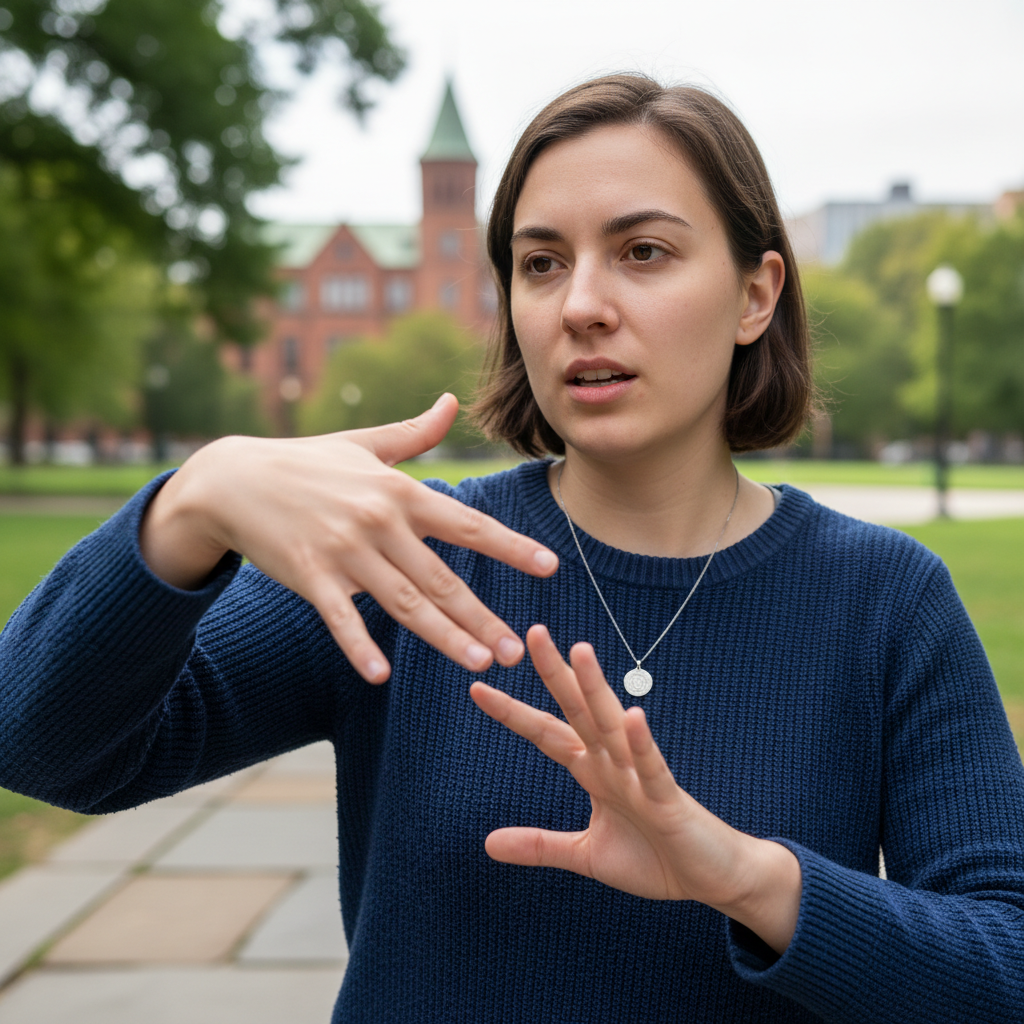} &
    \includegraphics[width=0.95in]{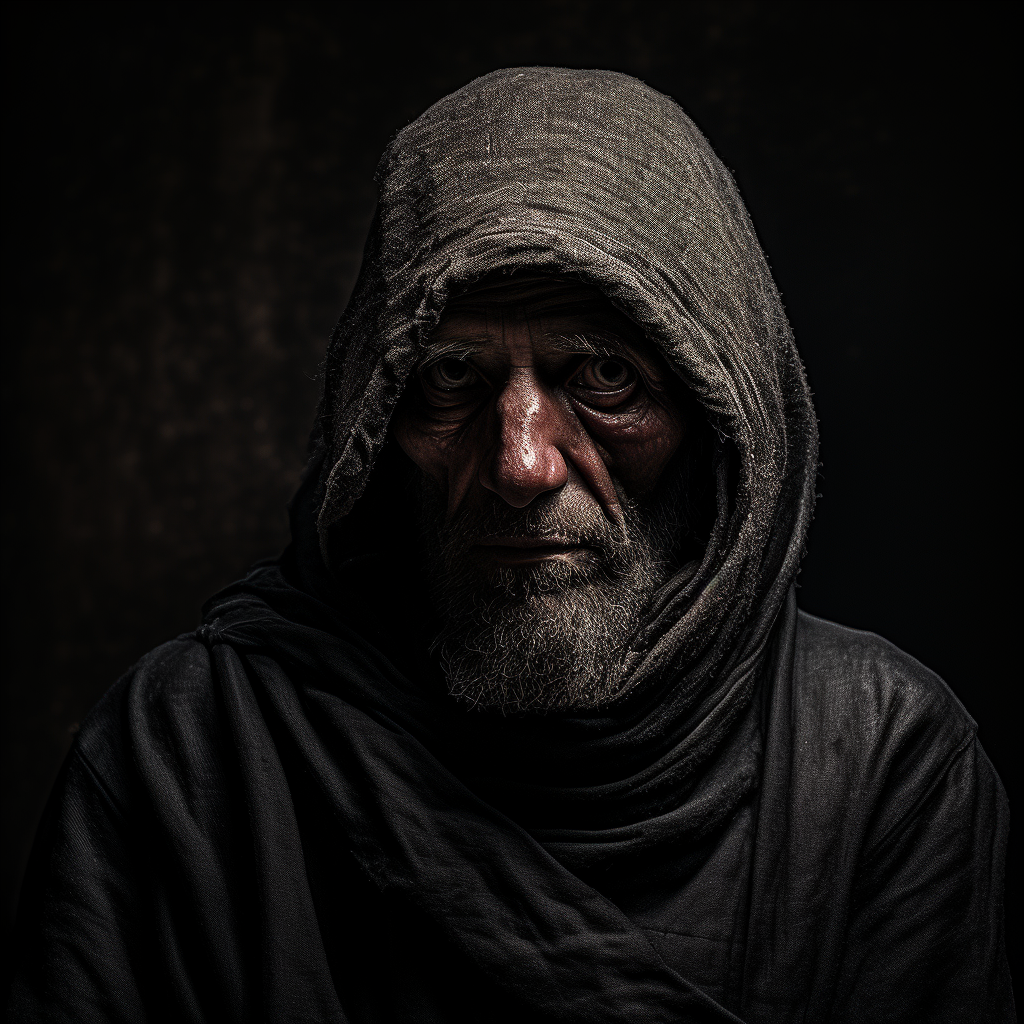} &
    \includegraphics[width=0.95in]{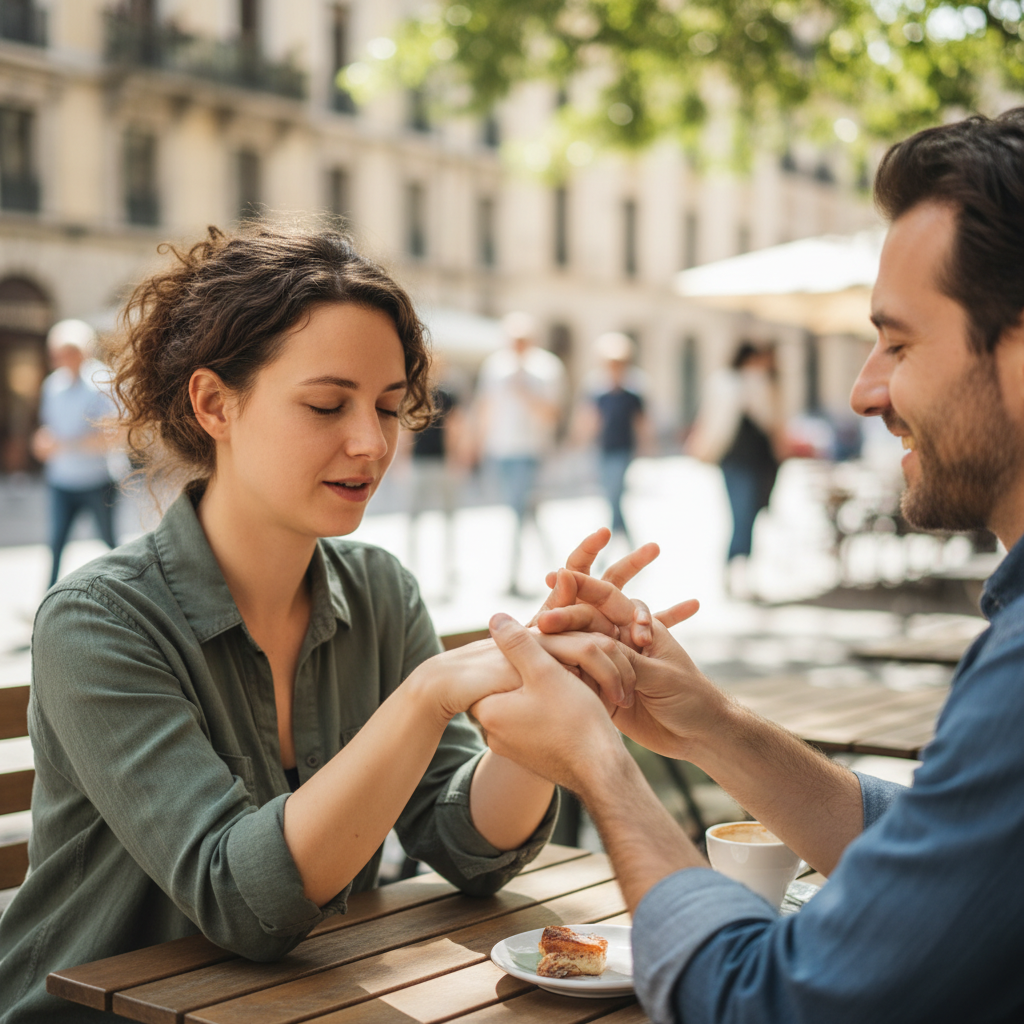} &
    \includegraphics[width=0.95in]{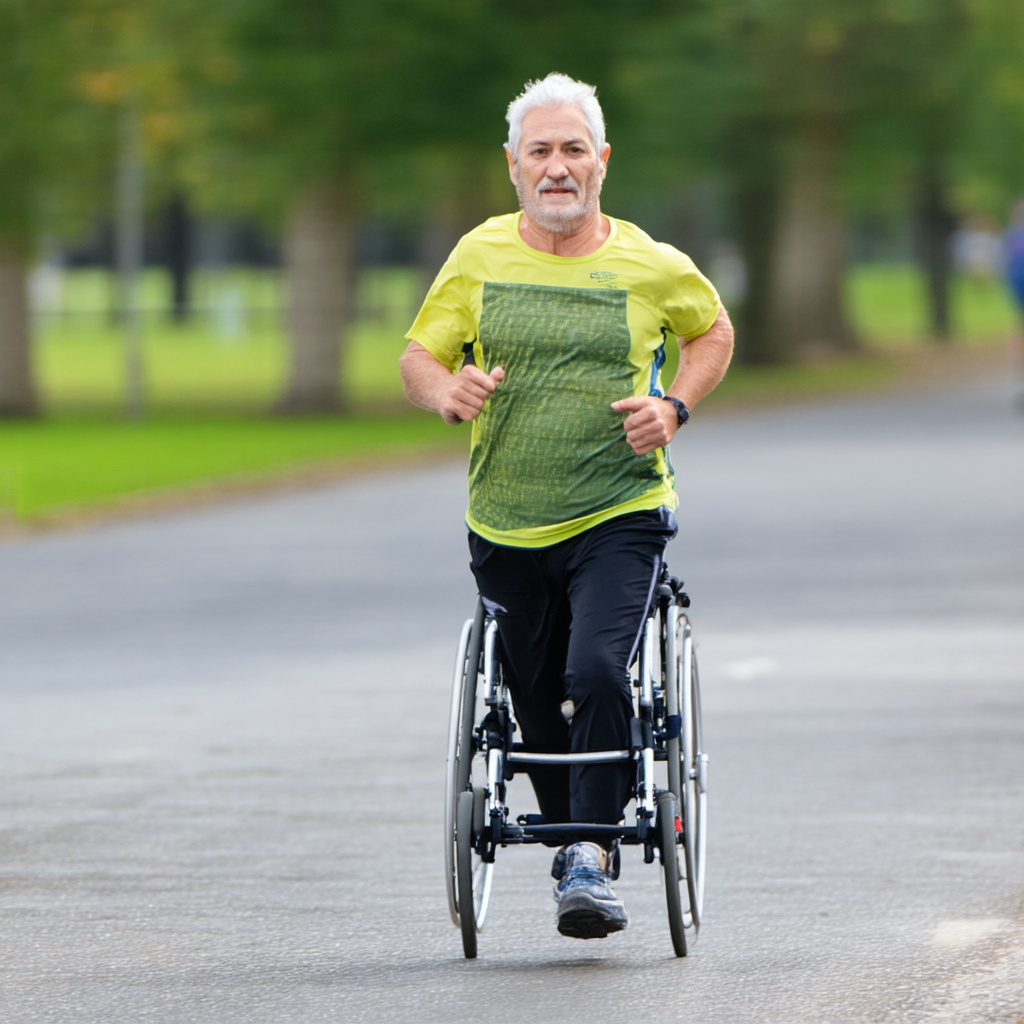} &
    \includegraphics[width=0.95in]{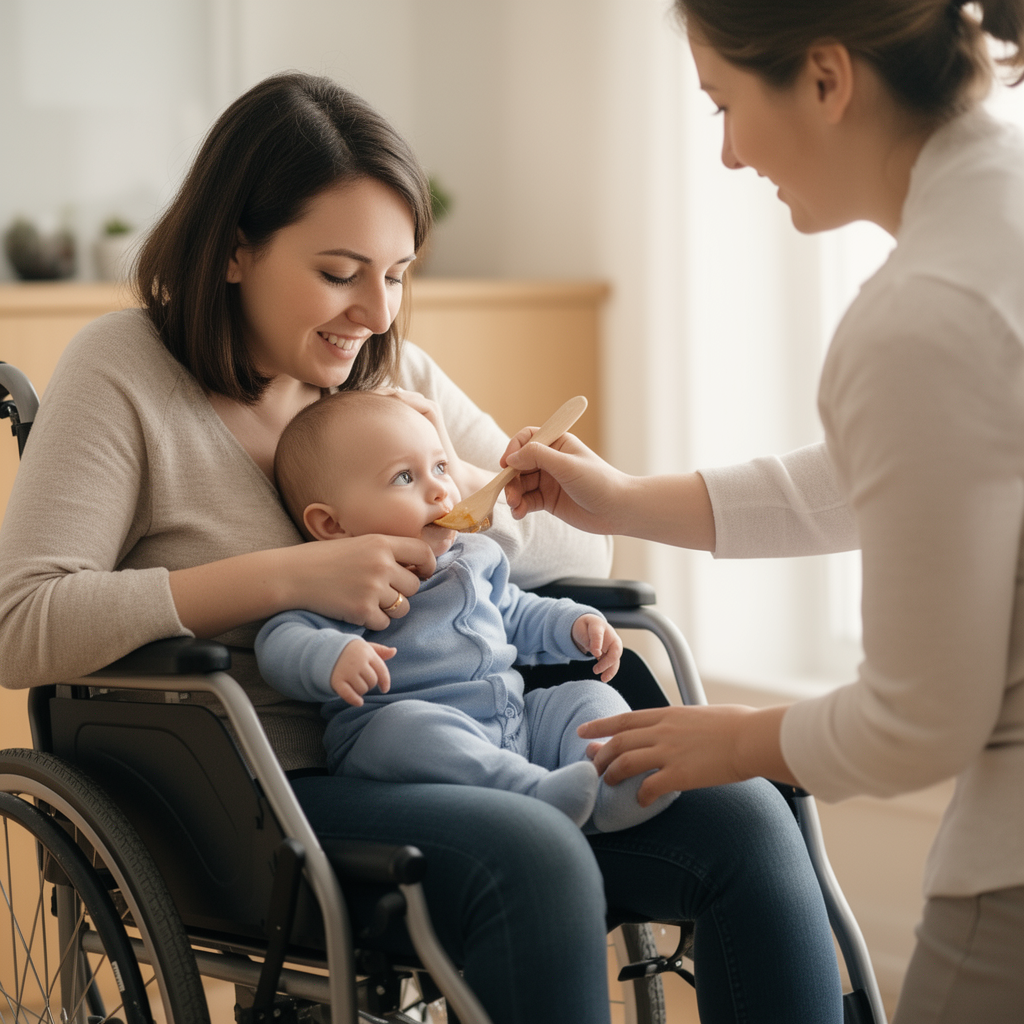} &
    \includegraphics[width=0.95in]{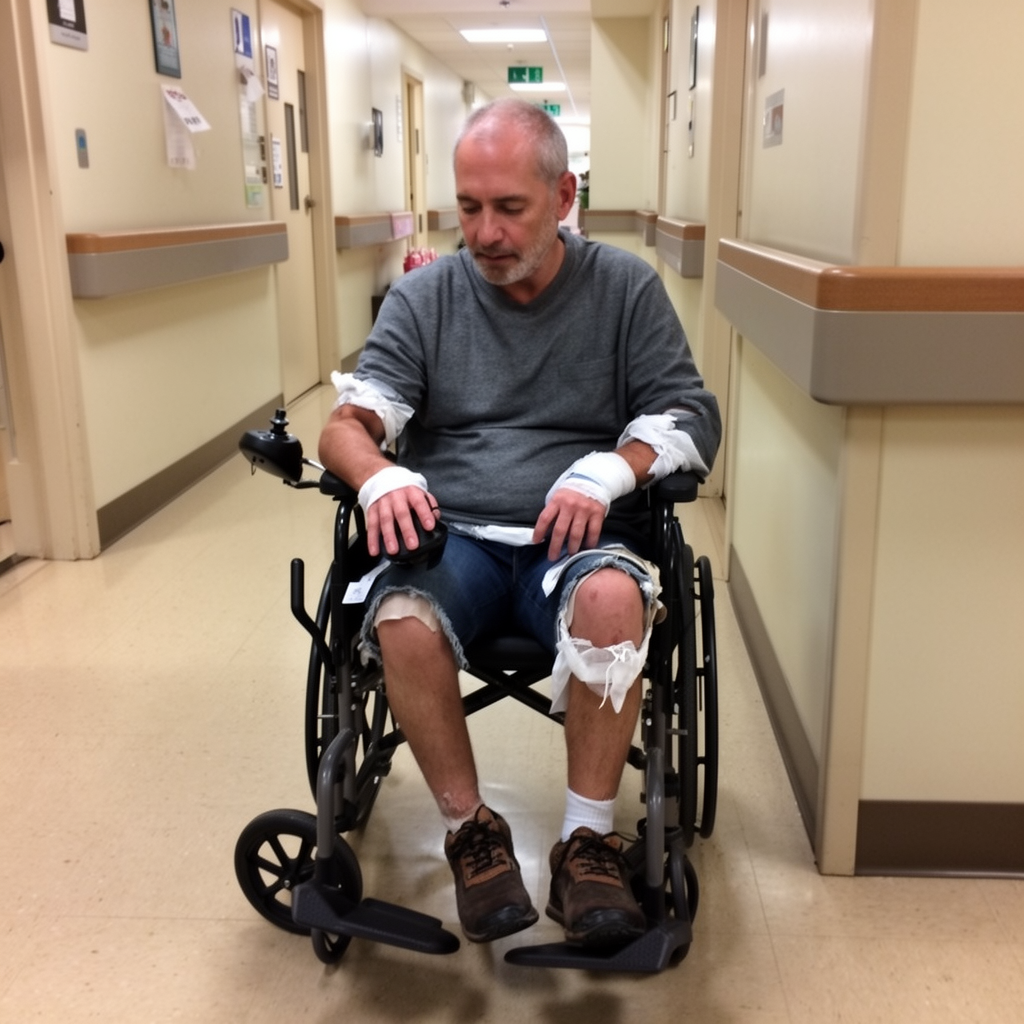} \\
  };
  
    \node[above=0.1cm of m-1-1.north] {\bfseries Blind};
    \node[above=0.1cm of m-1-2.north] {\bfseries Deaf};
    \node[above=0.1cm of m-1-3.north] {\bfseries Mute};
    \node[above=0.1cm of m-1-4.north] {\bfseries Deafblind};
    \node[above=0.1cm of m-1-5.north] {\bfseries Mobility};
    \node[above=0.1cm of m-1-6.north] {\bfseries Disabled};
    \node[above=0.1cm of m-1-7.north] {\bfseries Impaired};
\end{tikzpicture}    
\caption{Sample images from INCLUDE-BENCH across seven function groups categories: Blind, Deaf, Mute, Deafblind, Mobility impaired, Disabled, and Impaired. Images illustrate diverse contexts, activities, and demographic attributes generated from prompts in the benchmark.}
\label{fig:sample_images}
\end{figure*}

\begin{figure*}
    \centering
    \includegraphics[width=\linewidth]{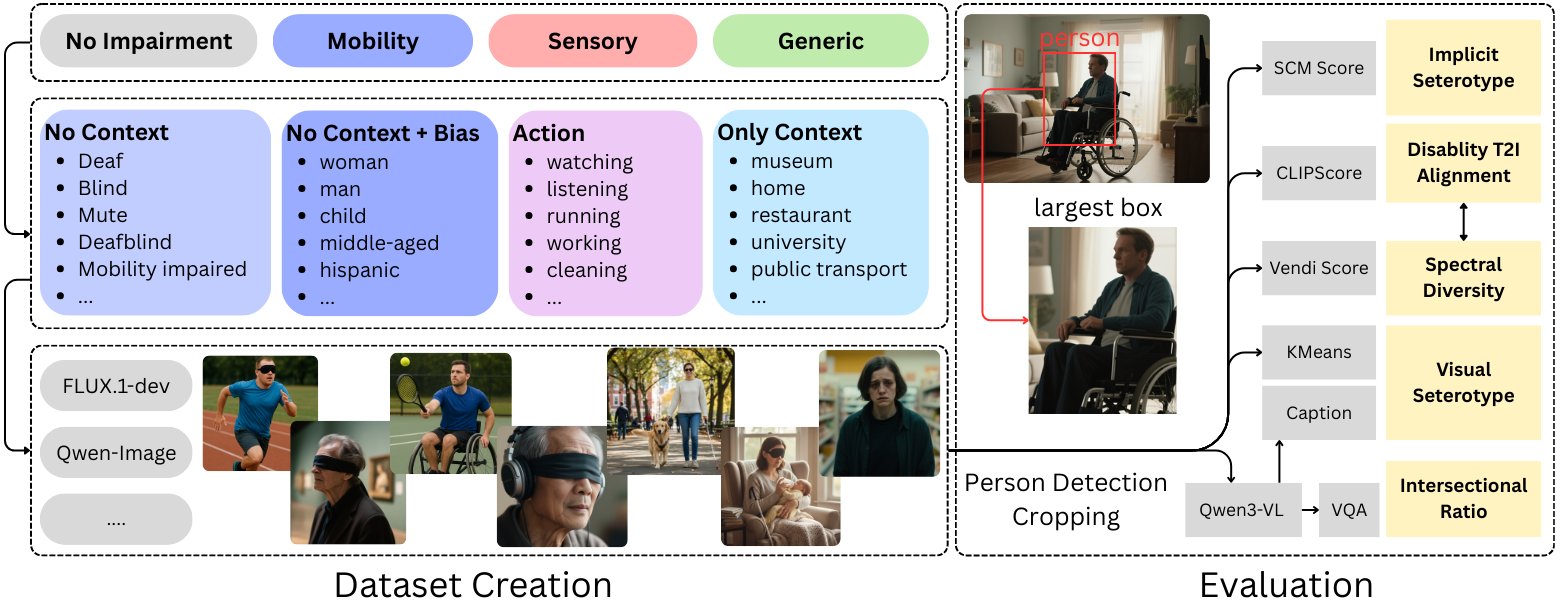}
    \caption{INCLUDE-BENCH pipeline: Dataset creation based on different disability groups across diverse context sets and image generation using 17 models. Evaluation involves constructing a person-centric dataset by cropping generated images and analyzing them using our evaluation toolbox for stereotype assessment}
    \label{fig:pipeline}
\end{figure*}

\noindent
Text-to-image (T2I) models, which synthesize images from textual descriptions, have rapidly advanced in both capability and popularity, and are increasingly used in digital content creation and other creative domains.
Despite these advances, T2I models often reproduce societal biases present in their training data. While gender, age, race, and cultural biases are relatively well studied \cite{wan2024surveybiastexttoimagegeneration, vazquez2024taxonomybiasesimagescreated}, other marginalized groups and intersectional identities remain underexplored. T2I models also raise concerns around toxicity, fairness, and authenticity, with representational harms that can stereotype, erase, or demean social groups, limiting individuals’ control over their depiction and contributing to negative cognitive and emotional outcomes \cite{Bavalatti_2025, blodgett2020language, blodgett2021stereotyping, katzman2023taxonomizing, shelby2023sociotechnical, wang2022measuring}.
\\
Disability representation has received limited attention, despite affecting a substantial portion of the population: an estimated 16\% globally \cite{WHO2022HealthEquityDisabilities}, and approximately 28.7\% of U.S. adults, with cognitive (13.8\%) and mobility (12.2\%) impairments most common \cite{cdc_disability_impacts_all_of_us}. Prevalence also varies by gender, with 20\% of women and 12\% of men living with a disability \cite{UNWomen2025DisabilityInclusion}. Yet representation remains minimal: a 2024 study found that only 2\% of respondents with disabilities felt represented in the media, indicating that accessible experiences, accurate depiction, and authentic narratives remain largely unmet \cite{ valuable500_inclusive_representation_2024}. Existing studies show that T2I models frequently produce stereotypical and narrow portrayals of persons with disability (PWD), often depicting them as older, sad and reliant on old-fashioned aids (e.g. manually operated wheelchairs). Community-centered analyses further reveal inaccurate, unsafe, and homogeneous depictions, persistent reliance on wheelchair tropes, erasure of multifaceted identities, and failures to represent certain type of disablity \cite{tevissen2024disability, bennett2025toward, mack2024}.
\\
In sociology, stereotypes are defined as generalized beliefs about social groups, ``mental representations of real differences between groups'', that simplify information processing \cite{bordalo2016stereotypes}. The representativeness framework \cite{bordalo2016stereotypes} states that stereotypes arise from selective recall of the most distinctive, diagnostic features, which maximize between-group differences and exhibit low within-group variation \cite{hilton1996stereotypes}, often compensating for limited information. From this perspective, T2I models over-weight highly diagnostic visual features, such as wheelchairs, when generating images from disability-related prompts, resulting in stereotypical and narrow portrayals.
\\
This paper presents INCLUDE-BENCH, a large-scale, comprehensive benchmark for evaluating disability bias in T2I models across diverse static and dynamic contexts. INCLUDE-BENCH addresses the lack of contextual evaluation in prior work, capturing activities and everyday realistic settings that play a crucial role in evaluating stereotypical perceptions and associations of disability. The benchmark also incorporates intersectional \cite{crenshaw2013mapping} representation along the dimensions of age, race, and gender.  Figure~\ref{fig:pipeline} provides an overview of the pipeline used to construct and evaluate the dataset. Our contributions are as follows:
(1) We construct INCLUDE-BENCH by curating multiple disability function group (generic, mobility, sensory) prompts across four subsets to probe T2I models for disability representation bias in a differential manner relative to non-disability prompts. 
(2) For each prompt, we generate 20 images across 17 state-of-the-art T2I models, resulting in a total of 119,680 images (Sample Images are shown in Figure \ref{fig:sample_images}).
(3) We show that mobility-impaired and default disability prompts consistently produce wheelchair depictions across all models. Disability-conditioned generations exhibit reduced diversity, and stereotypical portrayals demonstrate significantly stronger alignment with textual prompts that explicitly mention disability.
(4) We introduce a multimodal Stereotypical Semantic Aligment based on the Stereotype Content Model (SCM; \cite{granjon2024disability}), which reveals that T2I models systematically reproduce real-world disability stereotypes.


\section{Related Works}
\noindent
Generative AI reproduces and amplifies existing social biases in its outputs \cite{bennett2021s, bianchi2023easily, bommasani2021opportunities, dev2021harms, gadiraju2023wouldn, garg2018word, kiritchenko2018examining, qadri2023ai, shelby2023sociotechnical, weidinger2022taxonomy}, raising concerns about potential harms to marginalized groups \cite{bianchi2023easily, gadiraju2023wouldn, qadri2023ai, prabhakaran2020participatory, suresh2022towards}. Historically, PWD have been portrayed reductively or fantastically, as ``exotic'', ``freaks''; inspiring heroes, or dependent on charity \cite{garland2002politics, ellcessor2017disability, quayson2007aesthetic}. Realistic, identity-aligned depictions are crucial for normalizing respectful narratives, and research shows that PWD prefer visual representations, including avatars, that reflect their lived experiences \cite{garland2020integrating, garland2011misfits, garland2005feminist, garland2005disability, mack2024, zhang2022s}. Community-centered approaches further emphasize participatory evaluation to ensure AI aligns with disabled users’ perspectives \cite{brewer2023envisioning, gadiraju2023wouldn}.\\
Despite this, generative AI often misrepresents disability, producing micro-ableist stereotypes, biased classifications, and content misaligned with PWD lived experiences \cite{gadiraju2023wouldn, gogolushko2022ai, hutchinson2020unintended, venkit2025study, venkit2023automated, heung2024vulnerable, phutane2025cold, phutane2025disability, diaz2018addressing, herold2022applying, urbina2025disability, hassan2021unpacking, mei2023bias}. Standard evaluation metrics frequently fail to detect these harms, revealing bias against blind and low-vision users and disproportionate impacts on intersecting marginalized identities \cite{kapur-kreiss-2024-reference, phutane2025ableistintersectionaldisabilitybias}, especially in domains such as healthcare and finance \cite{panda2025accessevalbenchmarkingdisabilitybias}.\\
In the specific case of T2I models, disability representation remains comparatively underexamined \cite{cho2023dall, bianchi2023easily, naik2023social, trewin2018ai, whittaker2019disability}. Prior work shows that T2I outputs systematically misrepresent PWD, predominantly depicting older, sad individuals using manual wheelchairs, while women are shown younger and happier \cite{tevissen2024disability, bennett2025toward, mack2024}. Community-centered analyses reveal inaccurate, unsafe, and homogeneous depictions, persistent reliance on wheelchair tropes, erasure of multifaceted identities, and failures to generate certain conditions \cite{bennett2025toward}. Intersectional biases across age, race, and gender are common, and outputs often rely on emotional or sensationalized framings that reduce disability to narrow stereotypes \cite{bennett2025toward}.
Current social bias benchmarks for T2I models largely ignore disability, focusing more on race, gender, age, and cultural contexts in professions, skills, objects, or scenarios \cite{shukla2025biasconnectinvestigatingbiasinteractions, dincà2024openbiasopensetbiasdetection, girrbach2025large, dehdashtian2025oasis, chinchure2024tibetidentifyingevaluatingbiases, luo2025faintbenchholisticprecisebenchmark, luo2025bigbenchunifiedbenchmarkevaluating, lee2023holisticevaluationtexttoimagemodels, bakr2023hrsbenchholisticreliablescalable, wang2023t2iatmeasuringvalencestereotypical, kärkkäinen2019fairfacefaceattributedataset, luccioni2023stablebiasanalyzingsocietal, cho2023dallevalprobingreasoningskills, bansal2022texttoimagegenerativemodelsunderstand, liu2023equitablerepresentationtexttoimagesynthesis, jha2024visageglobalscaleanalysisvisual, huang2024kittenknowledgeintensiveevaluationimage, kannen2025aestheticsculturalcompetencetexttoimage}. Benchmarks such as TIBET \cite{chinchure2024tibetidentifyingevaluatingbiases} and BiasConnect \cite{shukla2025biasconnectinvestigatingbiasinteractions} partially include disability but lack depth, limiting evaluation of disability-related stereotypes. Previous Benchmarks also reveal pronounced gender bias in actions, attributes, roles, ideology, and emotions \cite{porikli2025hiddenbiasmachinestereotypes, girrbach2025large}, however, such biases have not yet been examined in the context of PWD. 
    \begin{table*}[h!]
        \centering
        \begin{tabular}{l l l l}
            \toprule
            Subset & Prompt & Context \\
            \midrule
            \midrule
            Baseline 
                & ``Photo of a Person'' & -\\
            \midrule
            No Context 
                & ``Photo of a disabled Person'' & -\\
            \midrule
            \multirow{3}{*}{No Context+Bias } 
                &  ``Photo of a disabled woman'' & gender\\
                &  ``Photo of a disabled child'' & age\\
                &  ``Photo of a disabled asian person'' & race\\
            \midrule
            \multirow{3}{*}{Action} 
                &  ``Photo of a disabled Person reading a book'' & learning/knowledge \\
                &  ``Photo of a disabled Person chatting'' & communication \\
                &  ``Photo of a disabled Person walking'' & mobility \\
            \midrule
            Only Context 
                &  ``Photo of a disabled Person at a museum'' & public\\
            \bottomrule
        \end{tabular}
        \caption{Example Prompts from each subset}
        \label{tab:prompt_examples}
    \end{table*}

\section{INCLUDE-BENCH}

    \subsection{Dataset Construction}        
        \noindent Following Garland-Thomson’s framing of disability as ``a relationship between bodies and their environments'' \cite{garland2011misfits}, we construct a prompt dataset to systematically probe how T2I models represent PWD in different environments. By designing prompts that vary key factors independently, we can isolate the effects of impairment, intersectional identity, and context, enabling controlled evaluation of representational biases.
        Prompts vary along three axes: (i) impairment specificity (none, generic, mobility and sensory), (ii) intersectional attributes (gender, race, age), and (iii) environmental context (none, static, dynamic). The dataset contains 352 prompts, organized into four subsets: \textbf{No Context} (8 prompts), \textbf{No Context + Bias} (96 prompts), \textbf{Only Context} (96 prompts), and \textbf{Action} (152 prompts grounded in the WHO ICF\footnote{International Classification of Functioning, Disability, and Health -  Activities and Participation \url{https://apps.who.int/classifications/icfbrowser/}}\cite{who_icf}). Example prompts for each subset are shown in Table \ref{tab:prompt_examples}. Each subset includes baseline, generic disability ($G \in \{\text{Disabled, Impaired}\}$), and impairment-specific conditions. The latter can be one of the following:      
        $F \in \{\text{Mute, Deaf, Mobility Impaired, Deafblind, Blind}\}$. We focus on impairments with observable visual characteristics, excluding cognitive and intellectual disabilities which are less visually apparent.
        The \textbf{No Context} subset tests baseline representations without environmental cues. The \textbf{No Context + Bias} subset includes intersectional variants (e.g. hispanic, middle-eastern). The \textbf{Only Context} subset introduces static spatial private and public settings (e.g., museum, cafe, park, home) to examine how environment shifts representation. The \textbf{Action} subset evaluates agentic representations, using prompts based on ICF domains (e.g. mobility, communication, learning/applying knowledge, self-care, domestic life) to test whether models depict PWD in contextually coherent scenes.


    \subsection{Representation and Stereotype Evaluation}
    \noindent In this section, we evaluate representation and implicit stereotypes of PWD in T2I models using multiple metrics. We analyze visual representation by clustering generated images and examining captions to identify recurring patterns. We then assess the interplay between disability label alignment and image spectral diversity. For implicit stereotypes, we introduce the SCM Score, which links visual representations to a sociological framework of stereotyping. Finally, we evaluate intersectional representation by measuring differential identity ratios across age, race, and gender when disability is specified compared to when it is not.

            \begin{table*}
            \centering
            \begin{tabular}{p{0.5cm}p{0.5cm}p{0.5cm}|p{0.8cm}p{0.8cm}p{1.5cm}p{1.3cm}p{1.3cm}p{1.5cm}p{0.8cm}|p{0.9cm}}
            \toprule
            \multicolumn{3}{c}{Intersection} & Blind & Deaf & Deafblind & Disabled & Impaired & Mobility & Mute & Mean \\
            \midrule
            \midrule
            o & f & w & 5.45 & 3.16 & 11.90 & 11.40 & 4.84 & 19.62 & 0.92 & 8.19 \\
            o & m & w  & 7.19 & 2.91 & 4.39 & 7.50 & 8.03 & 9.37 & 1.83 & 5.89 \\
            o & m & a & 5.49 & 1.28 & 1.69 & 3.09 & 4.85 & 2.91 & 2.97 & 3.18 \\
            m & f & w & 1.69 & 2.16 & 5.40 & 1.88 & 1.35 & 2.68 & 0.73 & 2.27 \\
            m & m & w & 1.96 & 2.32 & 1.79 & 1.95 & 2.72 & 1.66 & 1.35 & 1.96 \\
            m & m & a & 1.90 & 2.02 & 1.23 & 1.59 & 1.80 & 1.25 & 2.05 & 1.69 \\
            y & m & w & 0.59 & 0.75 & 0.54 & 0.66 & 0.69 & 0.35 & 1.18 & 0.68 \\
            y & m & af & 0.59 & 0.83 & 0.54 & 0.40 & 0.49 & 0.23 & 1.09 & 0.60 \\
            y & f & w & 0.36 & 0.72 & 0.41 & 0.31 & 0.22 & 0.22 & 0.56 & 0.40 \\
            y & m & a & 0.39 & 0.41 & 0.28 & 0.28 & 0.26 & 0.10 & 0.76 & 0.35 \\
            \bottomrule
            \end{tabular}
            \caption{Intersectional Identity in VQA (age o=old m=middle-aged and y=young; gender f=female and m=male; race w=white, a=asian and af=african): Ratios $R_{i,g}$ of Impairment group divided by No Impairment Across most common Intersections}
            \label{tab:intersection_ratio}
        \end{table*}

        \subsubsection{Stereotypical Visual Representations}
            \noindent To construct a person-centric evaluation set, we apply SAM3\footnote{\url{https://huggingface.co/facebook/sam3}} \cite{carion2025sam3segmentconcepts}, following \cite{girrbach2025large}, on all generated images and retain only detections labeled as \emph{person}. When multiple persons are detected, we select the largest bounding box. The resulting regions are cropped to create a person-focused dataset for representation analysis, and images without valid detections (108 in total) are excluded. This step reduces background confounds and ensures analyses focus on the visual depiction of individuals rather than scene composition.
            Using this dataset, we cluster all images into 10 clusters with MiniBatchKMeans\cite{sculley2010web} to capture common visual patterns and examine how these patterns align with the corresponding functional groups. We use Qwen3-VL-8B-Instruct\footnote{\url{https://huggingface.co/Qwen/Qwen3-VL-8B-Instruct}} \cite{bai2025qwen3vltechnicalreport} for image captioning on the person-centric dataset. The resulting captions provide semantic descriptions for each image, which we use to characterize clusters and identify shared visual cues via token frequency analysis. We further utilize the model for visual question answering (VQA) to extract demographic attributes (race, age, and gender) for intersectional analysis (See Supplementary Material). For each intersectional identity combination, we compute its relative appearance ratio by dividing its frequency within each functional group by its corresponding frequency in the non-impairment baseline. This allows us to assess which identities are overrepresented and which are underrepresented relative to the baseline distribution.
            $$R_{i,g} = \frac{f_{i,g}}{f_{i,\text{baseline}}}$$
            where $f_{i,g}$ denotes the frequency of identity $i$ within group $g$, and $f_{i,\text{baseline}}$ denotes the frequency of the same identity in the non-impairment baseline.
    
        \subsubsection{Disability Semantic Alignment}
            Let $F$ denote a functional group and $G$ a generic disability label. We construct a disability-conditioned prompt $T_{F+G}$ (e.g., ``photo of a $F$ person''). For each image, we get CLIP\footnote{\url{https://huggingface.co/openai/clip-vit-large-patch14}}\cite{radford2021learning} image embeddings $\mathbf{I}_i$ and a text embedding $\mathbf{T}$ for $T_{F+G}$. The semantic alignment of each image with the disability-conditioned prompt is measured using the CLIPScore \cite{hessel2022clipscorereferencefreeevaluationmetric}:
            \[
            \text{CLIPScore}(I_i, T_{F+G}) = \mathbf{I}_i^\top \mathbf{T}.
            \]
            Higher CLIPScores indicate stronger semantic alignment between the image and the disability label.
    
        \subsubsection{Spectral Image Diversity}
            Given a prompt $T$ and a set of generated images $\{I_i\}_{i=1}^n$, we first get image embeddings using CLIP and DINO\footnote{\url{https://huggingface.co/facebook/dinov3-vitl16-pretrain-lvd1689m}} \cite{simeoni2025dinov3}. We then define the cosine similarity kernel matrix $K_{ij} = \mathbf{I}_i^\top \mathbf{I}_j, \quad \mathbf{K} \in \mathbb{R}^{n \times n}$. Let $\lambda_1,\dots,\lambda_n$ be the eigenvalues of the normalized kernel $\mathbf{K}/n$. The \emph{Vendi Score} (VS) \cite{friedman2022vendi} is defined as:
            \[VS_T(\mathbf{I}_1,\ldots,\mathbf{I}_n) = \exp\Bigg(-\sum_{i=1}^{n} \lambda_i \log \lambda_i \Bigg),\]
            We continue in the following with Vendi Score based on CLIP (See supplementary material for DINO). The Vendi Score quantifies the diversity of generated images for a given prompt. Higher scores indicate more diverse unique elements in a sample, while lower scores reflect higher identical similarity. It represents an entropy-based estimate of the number of distinct visual modes in the generated images for each prompt.

        \subsubsection{Implicit Stereotypical Semantic Alignment}
            The Stereotype Content Model (SCM) \cite{fiske2018model, granjon2024disability, rohmer2018implicit} is a common framework to conceptualize perceptions of social groups along two dimensions: \emph{warmth} (sociability, morality) and \emph{competence} (ability, agency). PWD are often perceived as high in warmth but low in competence, reinforcing notions of vulnerability and lower social status \cite{sadler2012stereotypes, sadler2015competence, granjon2024disability}. 
            SCM links perceived social structures to stereotypes, which in turn predict emotional prejudices (e.g., pride, pity, contempt) and discriminatory behaviors \cite{cuddy2007bias}. 
            Let $\{I_i\}_{i=1}^{N}$ be images generated from a prompt $T$, with $\mathbf{I}_i$ denoting their CLIP embeddings. Let $\{\mathbf{z}_{w^+}^{(r)}\},\{\mathbf{z}_{w^-}^{(r)}\}$ and $\{\mathbf{z}_{c^+}^{(r)}\},\{\mathbf{z}_{c^-}^{(r)}\}$ be embeddings of positive and negative textual prompts, adapted from \cite{fraser-etal-2021-understanding}, for warmth (e.g. trustworthy, friendly vs. cold, selfish) and competence (e.g. confident, smart vs. lazy, ignorant), respectively (See Supplementary Material for details). Define the mean attribute embeddings    
            \[ 
                \bar{\mathbf{z}}_{w^\pm} = \frac{1}{R}\sum_{r=1}^{R} \mathbf{z}_{w^\pm}^{(r)}, 
                \quad
                \bar{\mathbf{z}}_{c^\pm} = \frac{1}{R}\sum_{r=1}^{R} \mathbf{z}_{c^\pm}^{(r)},
            \]
            and the normalized semantic directions
            \[
            \boldsymbol{\delta}_W = \frac{\bar{\mathbf{z}}_{w^+}-\bar{\mathbf{z}}_{w^-}}{\|\bar{\mathbf{z}}_{w^+}-\bar{\mathbf{z}}_{w^-}\|_2}, 
            \quad
            \boldsymbol{\delta}_C = \frac{\bar{\mathbf{z}}_{c^+}-\bar{\mathbf{z}}_{c^-}}{\|\bar{\mathbf{z}}_{c^+}-\bar{\mathbf{z}}_{c^-}\|_2}.
            \]
            The SCM score for each image is     
            \[
                \mathrm{SCM\_Score}_{\mathrm{W}}(I_i) = \mathbf{I}_i \cdot \boldsymbol{\delta}_W, 
                \quad
                \mathrm{SCM\_Score}_{\mathrm{C}}(I_i) = \mathbf{I}_i \cdot \boldsymbol{\delta}_C.
            \]  
            Higher SCM scores indicate stronger alignment with positive perceptions of warmth or competence. In our context, this allows quantifying whether disability-conditioned prompts systematically generate images consistent with societal stereotypes.

        \begin{table*}[]
            \centering
            \begin{tabular}{p{1.3cm} p{1cm} p{1cm} p{3.5cm} p{1.8cm} p{3cm}}
            \toprule
            ID (\%) & CLIP & Vendi & Function Group (\%) & Common & Context \\
             & Score &  &   & Terms &  \\
            \midrule
            \midrule
            \makecell[l]{1 (12.1)} 
                & 0.24 & 1.96 
                & \makecell[l]{Mobility (47)\\Disabled (43)\\IMP (8)} 
                & \makecell[l]{wheelchair \\man \\elderly \\neutral} 
                & \makecell[l]{\textbf{Public/Mobility}: \\commuting, park, \\sports, garden, pool, \\university, museum} \\
            \midrule
            \makecell[l]{2 (10.4) } 
                & 0.22 & 1.96 
                & \makecell[l]{Mobility (43)\\Disabled (38)\\IMP (9)} 
                & \makecell[l]{wheelchair \\woman \\elderly \\smiling} 
                & \makecell[l]{\textbf{Domestic/Social}: \\home, reading, cafe\\ voting, cleaning, \\ communication} \\
            \midrule
            \makecell[l]{3 (13.6) } 
                & 0.20 & 2.69 
                & \makecell[l]{Blind (32) DB (22)\\IMP (13) Mute (13) \\ Deaf (13)} 
                & \makecell[l]{man \\neutral \\dark \\ blindfold} 
                & \makecell[l]{\textbf{Public/Mobility}: \\ voting, public \\ transport, museum, \\restaurant} \\
            \midrule
            \makecell[l]{4 (11.2)} 
                & 0.17 & 2.42 
                & \makecell[l]{Deaf (22) NI (19)\\ Mute (17)  IMP (13) \\ DB (12) Blind (12)} 
                & \makecell[l]{man\\ young\\ hands} 
                & \makecell[l]{\textbf{Cognitive}: \\watching, listening,\\ working,\\ communication} \\
            \midrule
            \makecell[l]{5 (10.0)} 
                & 0.19 & 2.25 
                & \makecell[l]{IMP (24) Blind (18) \\ Mute (16)  DB (16) \\ Deaf (11)} 
                & \makecell[l]{elderly\\ man\\ neutral} 
                & \makecell[l]{drinking tea, \\public transport,\\ home, garden} \\
            \midrule
            \makecell[l]{6 (9.1)} 
                & 0.20 & 2.86 
                & \makecell[l]{Mute (23)  NI (21)\\Blind (14) IMP (12) \\Deaf (10)} 
                & \makecell[l]{dark\\ man} 
                & \makecell[l]{\textbf{Public/Mobility}: \\ walking, museum, \\park, garden,\\ commuting, cleaning, \\ public transport} \\
            \midrule
            \makecell[l]{7 (14.8)} 
                & 0.19 & 2.85 
                & \makecell[l]{NI (26) Deaf (23)\\Mute (18)  DB (15)} 
                & \makecell[l]{dark\\ man\\ young} 
                & \makecell[l]{\textbf{Public/Social}: \\ university, restaurant, \\ cafe, chatting, \\garden, park} \\
            \midrule
            \makecell[l]{8 (7.7)} 
                & 0.16 & 2.22 
                & \makecell[l]{NI (17) DB (17) \\ Deaf (16) Mute (13) \\ IMP (13) Blind (12)} 
                & \makecell[l]{woman}
                & \makecell[l]{\textbf{Domestic}: \\cooking, feeding \\baby, eating, \\cleaning, grocery} \\
            \midrule
            \makecell[l]{9 (5.7)} 
                & 0.16 & 2.05 
                & \makecell[l]{NI (19) Deaf (16)\\DB (16) IMP (14)\\Mute (13) Blind (12)} 
                & \makecell[l]{man\\ young\\ glasses} & \makecell[l]{\textbf{Intellectual}: \\ studying, reading, \\ writing, working, \\ university, voting} \\
            \midrule
            \makecell[l]{10 (5.4)} 
                & 0.18 & 2.43 
                & \makecell[l]{NI (17) Deaf (16) \\ Blind (15) Mute (15) \\ IMP (14)  DB (14)} 
                & \makecell[l]{man} 
                & \makecell[l]{\textbf{Physical}: \\running, gym, \\sports, pool} \\
            \bottomrule
            \end{tabular}
            \caption{Clusters with percentage of total prompts in dataset, distribution of functional groups within clusters, most common caption tokens, and most common context (NI=No Impairment; DB=Deafblind; IMP=Impaired)}
            \label{tab:clusters}
        \end{table*}
        \begin{figure*}[h!tbp]
            \centering
            \begin{subfigure}[t]{0.45\textwidth}
                \centering
                \includegraphics[width=\textwidth]{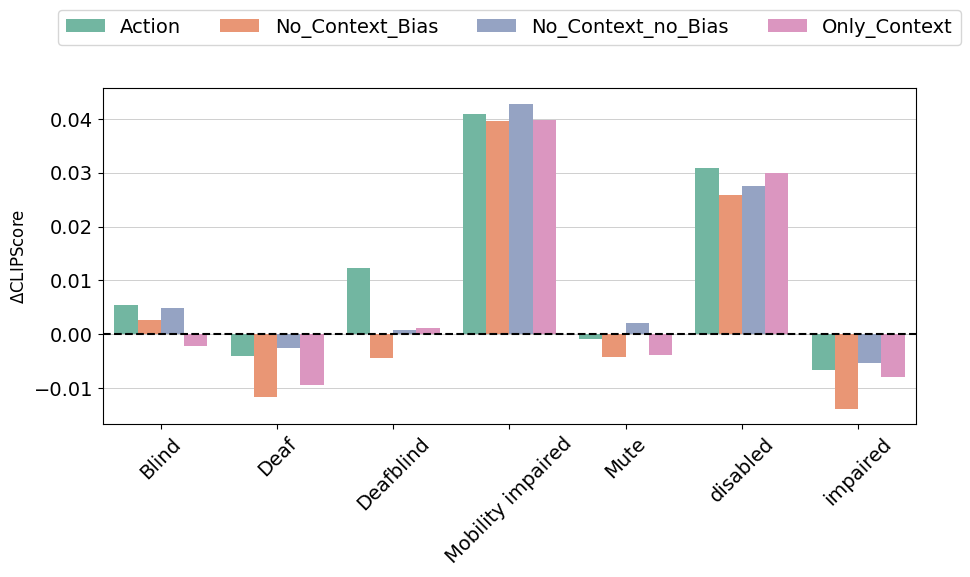}
                \caption{Difference $\Delta$ in ClipScore for each Impairment group based on No Impairment as baseline}
                \label{fig:ClipScore_Disability}
            \end{subfigure}
            \hfill
            \begin{subfigure}[t]{0.5\textwidth}
                \centering
                \includegraphics[width=\textwidth]{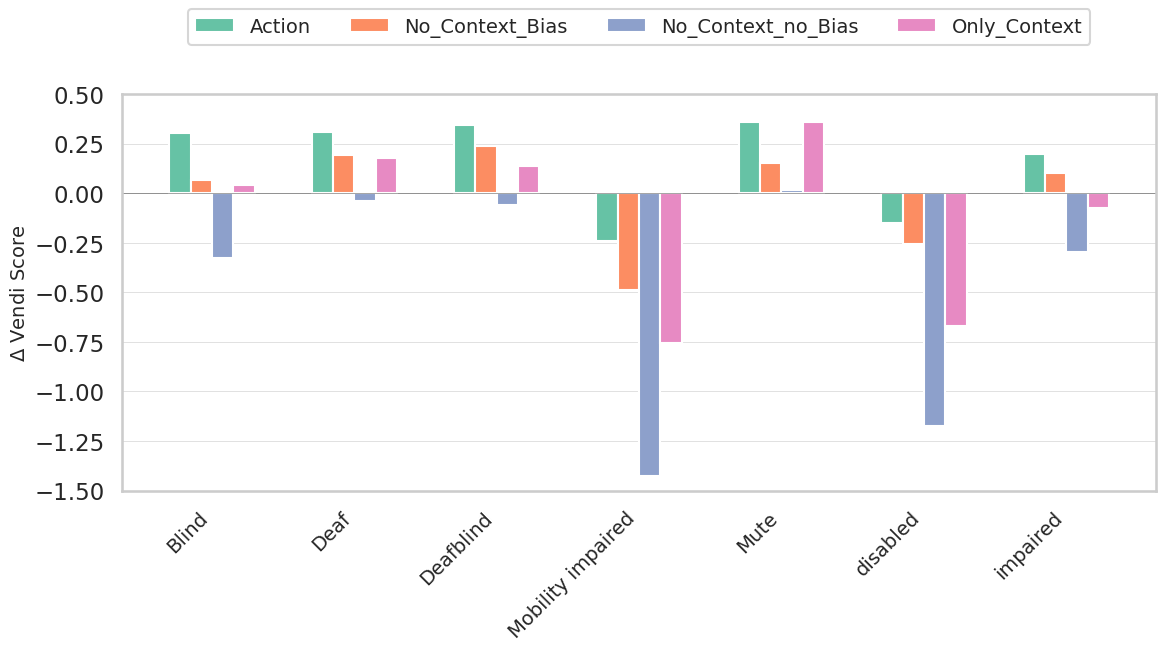}
                \caption{Difference $\Delta$ in Vendi Score for each Impairment group based on No Impairment as baseline}
                \label{fig:Vendi_delta_crop}
            \end{subfigure}
        \end{figure*}
        
\section{Characterizing Disability Depictions in T2I Models}        
    \noindent To assess representation across architectures and training paradigms, we evaluate a diverse set of T2I models, including the Stable Diffusion family\footnote{SD1.5\cite{rombach2022highresolutionimagesynthesislatent}, SDXL\cite{podell2023sdxlimprovinglatentdiffusion}, SD3.5-medium \cite{stable-diffusion-3.5-medium}, SD3.5-large\cite{stable-diffusion-3.5-large}}, the FLUX family\footnote{FLUX.1-schnell\cite{FLUX.1-schnell}, FLUX.1-dev\cite{FLUX.1-dev}, FLUX.2-dev \cite{FLUX.2-dev}}, as well as additional open models PixArt-$\alpha$ \cite{chen2023pixartalphafasttrainingdiffusion}, HiDream-l1 \cite{cai2025hidreami1highefficientimagegenerative}, CogView4-6B \cite{CogView4-6B}, Lumina2 \cite{qin2025luminaimage20unifiedefficient}, Playground-v2.5 \cite{li2024playgroundv25insightsenhancing}, SANA \cite{xie2024sanaefficienthighresolutionimage}, Qwen-Image \cite{wu2025qwenimagetechnicalreport}, and the unified architecture JanusPro-7B \cite{chen2025janusprounifiedmultimodalunderstanding}. We also evaluate the closed models, NanoBanana \cite{nanobanana} and GPT-Image-1-mini \cite{GPT-Image-1-mini}. For each prompt, 20 images are generated with random seeds using default inference parameters, resulting in 7,040 images per model and 119,680 images in total.

    \subsubsection{Stereotypical Visual Representations}        
        The clustering analysis reveals clear patterns of bias and representational shortcuts in how disabilities are depicted (Table \ref{tab:clusters}). Mobility impairments are predominantly shown through wheelchair imagery and strongly associated with elderly individuals, with men appearing in public or mobility-focused contexts (C1) and women largely confined to domestic settings (C2). Blindness-related clusters skew toward elderly men (C5) and often include explicit visual markers like blindfolds (C3), but also sunglasses, reinforcing symbolic and demographic stereotyping. 
        Sensory impairments, such as Deaf, Deafblind or Mute, are more consistently distinguished when tied to communication or cognitive tasks (C4), frequently emphasizing hands or gestures, and often portraying younger males. Gender roles are further differentiated across functional contexts for sensory disabilities, which are sometimes confused with no impairment. In particular, women appear primarily in domestic and caregiving settings (C 8), while men dominate public, intellectual, and physical activity contexts (C7, C9, and C10). 
        \\
        Across clusters, there is a consistent compression of age, segregation by gender roles, and visual simplification. Wheelchairs, blindfolds, and hand gestures dominate as visual shorthands, resulting in demographically homogeneous and narrow representations. 
        Clusters featuring wheelchair or blindfold depictions achieve the highest CLIPScores (C1-3), whereas clusters depicting sensory impairments in active or intellectual roles show lower CLIPScores (C10, 4, 9 and 8). Vendi diversity is lowest for wheelchair-heavy clusters (C1-2) and highest in clusters with more varied contexts (C6-7).
        \\
        Table \ref{tab:intersection_ratio} shows the results of the intersectional analysis, highlighting disparities in representation across age, gender, and race. Older white females consistently show the highest ratios across multiple disability types, particularly mobility impairment, disabled and deafblind, indicating they are disproportionately overrepresented in images generated in the disability conditions compared to baseline. White old white men are overrepresented in Blind and Impaired depictions.
        In contrast, younger individuals, which are more common in baseline, decrease the most. Middle-aged groups show moderate ratios, with slight gender differences, where middle-aged white women are more represented in deafblind and mobility impaired conditions and middle-aged white men in the impaired condition.
        Overall, this aligns with the findings in the cluster analysis, since it shows that older people with impairments are over-represented, and the gender bias showed that women are more associated in specific contexts such as domestic and caregiving  (reading, cooking, cleaning), while older white males are more often shown in active or public settings (cafe, running, sports) (See Supplementary  Material). This highlights intersectional biases in T2I models, where portrayals of PWD are influenced by contextual assumptions about age, gender, and race.
        

        \begin{figure}[h!tbp] 
            \centering
            \includegraphics[width=0.5\textwidth]{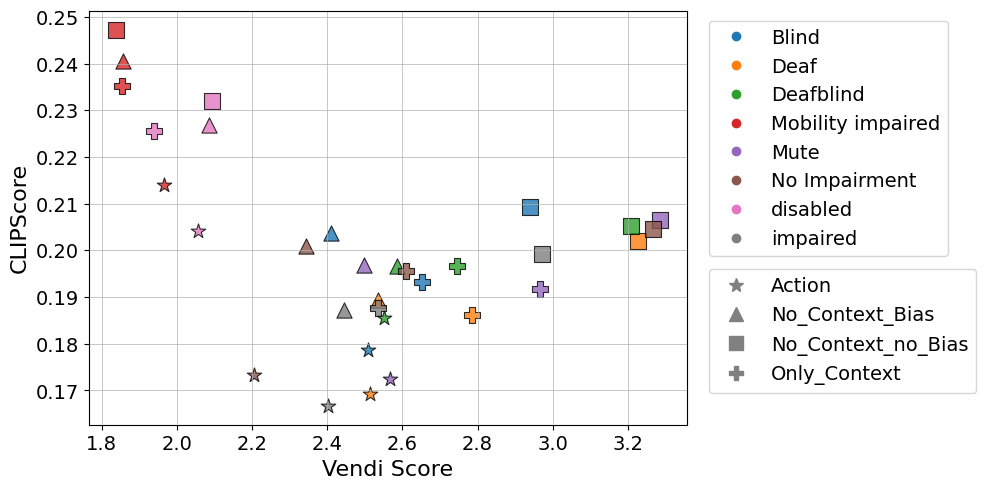}
            \caption{Vendi Score and CLIPScore for each Impairment group}
            \label{fig:Vendi_CLIP_Score}
        \end{figure}

    \subsubsection{Alignment-Diversity Tradeoff}
        Our results show a clear trade-off between alignment and diversity in T2I models (Figure \ref{fig:Vendi_CLIP_Score}). Physical and generic disabilities (Mobility Impaired, Disabled) achieve the highest CLIPScore (Figure \ref{fig:ClipScore_Disability}) but the lowest diversity (Figure \ref{fig:Vendi_delta_crop}), especially under No-Context and Only-Context settings, while sensory disabilities (Blind, Deaf, Mute, Deafblind) maintain moderate alignment and higher diversity, particularly in communicative and cognitive activities. Alignment is lowest in the Actions subset and highest in Baseline No-Context, whereas diversity decreases for physical tasks and increases for cognitive or social tasks (See supplementary material). This supports the findings of the cluster and intersectional analyses, showing that wheelchair depictions reduce representational diversity and semantically align certain demographics with stereotypical imagery.
        \\
        In the Actions subset, alignment is highest for cognitive and communicative tasks and lower for Domestic Life and Mobility. Mobility Impaired and Disabled show consistently high alignment but reduced diversity, while sensory groups exhibit lower alignment for physical activities but higher diversity, especially in messaging, commuting, sports, and work (Details about specific Contexts subsets and Models can be found in supplementary material). This demonstrates a task-dependent trade-off between alignment and diversity across disability types.

        \begin{figure}[htbp] 
            \centering
            \includegraphics[width=0.5\textwidth]{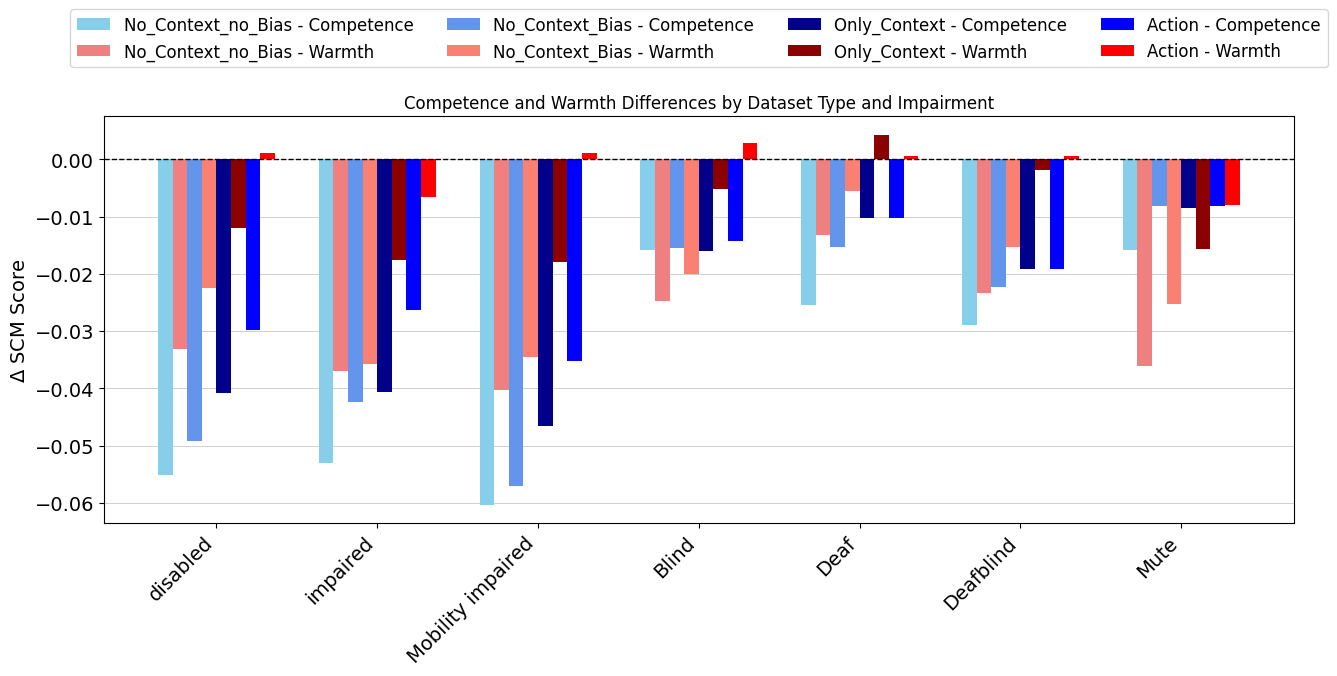}
            \caption{SCM Score for each Dataset and Impairment group}
            \label{fig:SCM}
        \end{figure}
        
    \subsubsection{Stereotypical Semantical Alignment}
\begin{figure}[htbp] 
            \centering
            \includegraphics[width=0.5\textwidth]{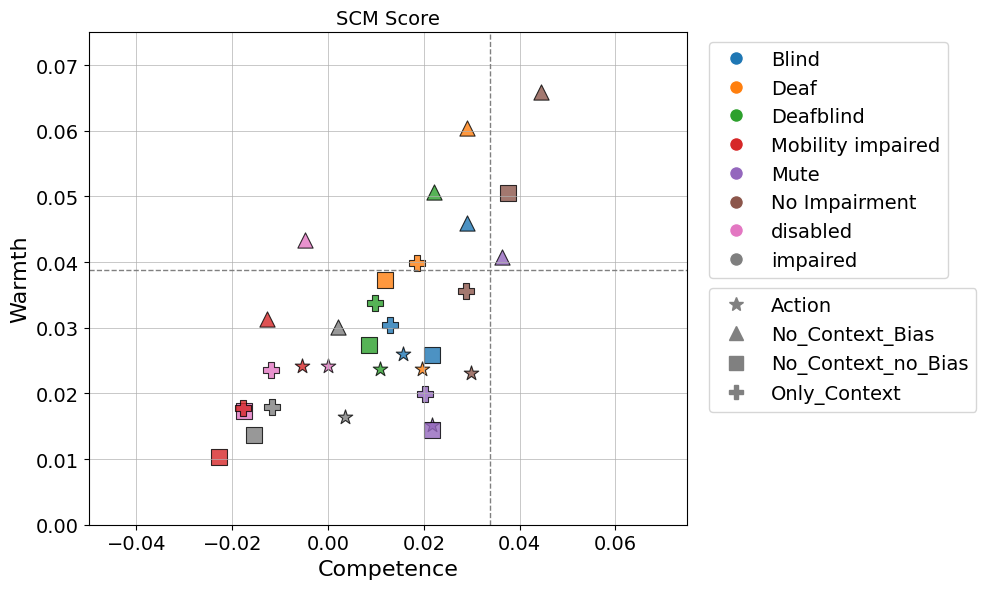}
            \caption{SCM Score for each Dataset and Impairment group}
            \label{fig:SCM_2dim}
        \end{figure}

        Figures \ref{fig:SCM} and \ref{fig:SCM_2dim} show mean SCM scores across subsets. Mobility Impaired, Disabled, and Impaired groups rank lowest on both warmth and competence, especially in the default (No-Context subset). Only-Context slightly improves scores, Action subset increases competence most, and No-Context-Bias mitigation primarily raises warmth. The No Impairment baseline shows highest competence but more variability in warmth. Sensory groups fall in between, with the Bias dataset yielding the largest warmth gains. Mute scores lowest on warmth, Deaf highest, and Blind and Deafblind in between.
        \\
        Images most associated with wheelchairs (mobility, disabled, impaired) are perceived as less warm and less competent than the non-impaired baseline. However, while static and dynamic context reduce perceived warmth for the baseline condition, they increase both warmth and competence for the mobility, disabled, and impaired conditions, possibly by signaling agency or reducing the salience of the disability cue. 
        Blind, Deaf, and Deafblind individuals show moderate decreases in competence, with larger reductions in tasks involving sensory or social engagement (e.g., reading a book, listening to someone, sending messages), but relatively smaller drops in physical or passive activities (See Supplementary material).
        In the literature, perceptions of PWD vary by disability type, with most groups rated higher on warmth than competence \cite{canton2023stereotype, sadler2012stereotypes, sadler2015competence, granjon2024disability}, whereas invisible disabilities are often perceived as less warm but more competent than visible ones \cite{granjon2024disability}. Our findings show that sensory (“invisible”) PWD are depicted with higher competence and, contrary to expectations, also higher warmth, compared to mobility-impaired and generic disability groups. Previous work on language models found that abled people are more strongly associated with both warmth and competence than PWD \cite{herold2022applying}, consistent with implicit human measures showing more negative associations on both dimensions for PWD \cite{rohmer2012implicit}, suggesting that social desirability may mask underlying biases in explicit evaluations. Our results mirror these patterns, non-disabled depictions are consistently rated higher on both warmth and competence than PWD.




\section{Conclusion}
\noindent This study introduced INCLUDE-BENCH, a novel benchmark for evaluating T2I models' representation of PWD and its intersections with other social variables. Our results provide a comprehensive evaluation of how contemporary T2I models represent disability. Across models, we observe a consistent trade-off between semantic alignment and visual diversity. Physical and generic disabilities, particularly Mobility Impaired and Disabled, achieve the highest CLIPScores but exhibit the lowest diversity, indicating reliance on narrow visual diagnostic features such as wheelchairs. Sensory disabilities show moderately higher diversity, yet remain constrained by recurring symbolic cues such as blindfolds or hand gestures. Contextual variation and bias mitigation strategies yield only limited improvements. They neither substantially expand representational diversity nor eliminate stereotypical compression. This pattern is consistent with sociological theories of stereotyping. T2I models tend to over-weight highly diagnostic visual markers (e.g., wheelchairs, blindfolds, hand gestures), generating images that maximize recognizability and label alignment at the expense of intra-group diversity.
\\
Intersectional analysis further reveals systematic demographic skew. Older white individuals, especially older white people, are disproportionately depicted as disabled, whereas younger and non-white individuals are underrepresented. Gender roles remain stratified across domestic and public contexts, and competence penalties are most pronounced for mobility-related impairments. These patterns extend beyond visual depiction to encode implicit social judgments about capability and agency.
\\
Overall, current T2I systems do not merely reflect disability, but reproduce simplified, demographically narrow, and socially stratified portrayals. The persistence of these patterns across models suggests that bias is structurally embedded.  These results underscore the importance of bias-aware design and evaluation, as well as the careful use of context, to achieve fairer and more inclusive visual representations of PWD.

\section{Limitations}
While SAM3, Qwen3-VL, and CLIP provide large-scale multimodal understanding for our evaluation, their use in INCLUDE-BENCH has several limitations. These models inherit biases from their pretraining data, which can amplify skewed associations in captions, VQA and embeddings. Our benchmark does not incorporate human annotation by PWD due to its scale. Reliance on automated metrics also limits capturing subjective harms, microaggressions, or culturally contextualized stereotypes. Furthermore, not all functional disability groups or assistive technologies are incorporated, and contexts remain limited. Future work will involve PWD-centered evaluation to ensure findings align with PWDs’ perceptions of authentic and respectful representations.

{\small
\bibliographystyle{ieee_fullname}
\bibliography{egbib}
}

\end{document}